\documentclass{article}

\usepackage[preprint]{neurips_2024}

\setcitestyle{numbers,square,sort,comma}
\usepackage{amsmath}
\usepackage{amssymb}
\usepackage{mathtools}
\usepackage{amsthm}

\usepackage[utf8]{inputenc} %
\usepackage[T1]{fontenc}    %
\usepackage{hyperref}       %
\usepackage{url}            %
\usepackage{booktabs}       %
\usepackage{amsfonts}       %
\usepackage{nicefrac}       %
\usepackage{microtype}      %
\usepackage{xcolor}         %

\usepackage[capitalize,noabbrev]{cleveref}

\theoremstyle{plain}
\newtheorem{theorem}{Theorem}[section]
\newtheorem{proposition}[theorem]{Proposition}

\theoremstyle{definition}
\newtheorem{definition}[theorem]{Definition}

\theoremstyle{remark}

\newcommand*\dbar[1]{\overline{\overline{\lower0.2ex\hbox{$#1$}}}}
\newcommand{\harrow}[1]{\mathstrut\mkern2.5mu#1\mkern-11mu\raise1.6ex%
  \hbox{$\scriptscriptstyle\rightharpoonup$}}
\newcommand{\ftheta}{\boldsymbol{\theta}^{(f)}}

\usepackage{changepage,threeparttable} %
\usepackage{makecell, multirow}
\usepackage{amsmath}
\usepackage{booktabs}
\usepackage{enumitem}
\usepackage{subcaption}
\usepackage{adjustbox}

\definecolor{dark2green}{rgb}{0.1, 0.65, 0.3}
\definecolor{dark2orange}{rgb}{0.9, 0.4, 0.}
\definecolor{dark2purple}{rgb}{0.4, 0.4, 0.8}
\newcommand{\first}[1]{\textbf{\textcolor{dark2green}{#1}}}
\newcommand{\second}[1]{\textbf{\textcolor{dark2orange}{#1}}}
\newcommand{\third}[1]{\textbf{\textcolor{dark2purple}{#1}}}

\usepackage{xargs}
\usepackage{xspace}
\usepackage[colorinlistoftodos,prependcaption,textsize=tiny,disable]{todonotes}

\usepackage{algorithm}
\usepackage{algorithmic}

\usepackage{wrapfig}
\usepackage{graphbox}
\newcommand{\inlinefig}[2][10]{\protect\includegraphics[align=c, height=#1pt]{./figures/#2.pdf}}

\newcommand{\pe}{PE\xspace}
\newcommand{\lepe}{LE\xspace}

\newcommand{\model}{GECO\xspace}
\newcommand{\modelfull}{\underline{G}raph-\underline{E}nhanced \underline{C}ontextual \underline{O}perator\xspace}
\newcommand{\X}{\mathbf{X}}
\newcommand{\Y}{\mathbf{Y}}
\newcommand{\Adj}{\mathbf{A}}
\newcommand{\D}{\mathbf{D}}
\newcommand{\Per}{\mathbf{\pi}}

\newcommand{\bn}{{BatchNorm}\xspace}
\newcommand{\layernorm}{{LayerNorm}\xspace}
\newcommand{\propagate}{{Propagate}\xspace}
\newcommand{\permute}{{Permute}\xspace}
\newcommand{\samplephi}{{SamplePermutation}\xspace}
\newcommand{\posenc}{{GraphPositionalEncoder}\xspace}
\newcommand{\ffn}{{FFN}\xspace}
\newcommand{\lcb}{{Local Propagation Block}\xspace}
\newcommand{\gcb}{{Global Context Block}\xspace}
\newcommand{\lcba}{{LCB}\xspace}
\newcommand{\gcba}{{GCB}\xspace}
\newcommand{\E}{{\mathcal{E}}\xspace}
\newcommand{\R}{{\mathcal{R}}\xspace}

\newcommand{\train}{\mathcal{D}}

\newcommand{\Jloss}{J}
\def\sN{{\mathbb{N}}}
\def\sH{{\mathbb{H}}}
\def\sR{{\mathbb{R}}}
\def\sF{{\mathbb{F}}}
\def\sH{{\mathbb{H}}}

\def\vh{{\mathbf{h}}}
\def\vx{{\mathbf{x}}}
\def\vy{{\mathbf{y}}}
\def\vtheta{{\mathbf{\theta}}}
\def\vhid{{\vh}}
\def\rvs{{\mathbf{s}}}
\def\gB{{\mathcal{B}}}
\def\rmZ{{\mathbf{Z}}}
\def\rvs{{\mathbf{s}}}

\title{A Scalable and Effective Alternative to Graph Transformers}

\author{%
   Kaan Sancak\thanks{School of Computational Science and Engineering, Georgia Institute of Technology, Atlanta, GA, USA}
   \thanks{This work was partially done when the first author was a research scientist intern at Meta AI.} \\
    \texttt{balin@gatech.edu} \\
    \And
    Zhigang Hua\thanks{Meta AI}\\
    \texttt{zhua@meta.com} \\
    \And
    Jin Fang\footnotemark[2]\\
    \texttt{fangjin@meta.com} \\
    \And
    Yan Xie\footnotemark[2]\\
    \texttt{yanxie@meta.com} \\
    \And
    Andrey Malevich\footnotemark[2]\\
    \texttt{amelevich@meta.com} \\
    \And
    Bo Long\footnotemark[2]\\
    \texttt{bolong@meta.com} \\
    \And
    Muhammed Fatih Bal{\i}n\footnotemark[1]\\
    \texttt{balin@gatech.edu} \\
    \And
    \"Umit V. \c{C}ataly\"urek\thanks{Amazon Web Services. This publication describes work performed at the Georgia Institute of Technology and is not associated with AWS.}\protect\phantom{\footnotesize 1}\footnotemark[1]\\
    \texttt{umit@gatech.edu} \\
}

\begin{document}

\maketitle

\begin{abstract}
    Graph Neural Networks (GNNs) have shown impressive performance in graph representation learning,
    but they face challenges in capturing long-range dependencies due to their limited expressive power.
    To address this, Graph Transformers (GTs) were introduced, utilizing self-attention mechanism to effectively model pairwise node relationships.
    Despite their advantages, GTs suffer from quadratic complexity
    w.r.t. the number of nodes in the graph,
    hindering their applicability to large graphs.
    In this work, we present \modelfull (\model),
    a scalable and effective alternative to GTs that leverages neighborhood propagation and global convolutions to effectively capture local and global dependencies in quasiliniear time.
    Our study on synthetic datasets reveals that \model reaches $169\times$ speedup on a graph with 2M nodes w.r.t. optimized attention. 
    Further evaluations on diverse range of benchmarks showcase that \model
    scales to large graphs where traditional GTs often face memory and time limitations. 
    Notably, \model consistently achieves comparable
    or superior quality compared to baselines, improving the SOTA up to $4.5\%$,
    and offering a scalable and effective solution for large-scale graph learning.
\end{abstract}

\section{Introduction}
\label{section:intro}

Graph Neural Networks (GNNs) have been state-of-the-art (SOTA) models for graph representation learning showing superior
quality across different tasks spanning node, link, and graph level
prediction~\citep{gori2005,scarselli2009, kipf2017semisupervised, zhang2018link, zhang2018end}.
Despite their success, GNNs have fundamental limitations that affect their ability to capture long-range dependencies in graphs.
These dependencies refer to nodes needing to exchange information over
long distances effectively,
especially when the distribution of edges is not directly related to the task
or when there are missing edges in the graph~\citep{dwivedi2022long}.
This limitation can further lead to information over-squashing caused by repeated propagations within GNNs~\citep{li2018deeper, alon2020bottleneck, topping2022understanding}.

Graph Transformers (GTs)~\citep{dwivedi2020generalization, ying2021do, wu2021representing} were introduced to overcome the limitations of GNNs
by incorporating the self-attention mechanism~\citep{vaswani2017attention},
and achieved SOTA across various benchmarks.
GTs can model long-range dependencies by attending to potential neighbors among the entire set of nodes.
However, GTs suffer from quadratic complexity in contrast to the linear time and memory complexity inherent in GNNs.
This quadratic complexity stems from that each node needs to attend to every other node,
preventing GTs' widespread adoption in large-scale real-world scenarios.
As mini-batch sampling methods for GTs remain under-explored,
the primary application of GTs has been on smaller
datasets, such as molecular ones~\citep{freitas2021a, hu2021ogb, dwivedi2022long, dwivedi2023benchmarking}.
Consequently, exploring novel efficient and high-quality attention replacements
remains a crucial research direction to unlock the full potential of GTs for large-scale graphs.
Recently, global convolutional language models have emerged as promising alternatives for attention~\citep{romero2022ckconv,li2023what}.
Specifically, Hyena~\citep{poli2023hyena} has demonstrated impressive performance,
offering efficient processing of longer contexts with high quality.

In this work, we aim to find an efficient alternative to dense attention mechanisms
to scale graph transformers without sacrificing the modeling quality.
The key challenge lies in designing an efficient model that can effectively capture both local and long-range dependencies within large graphs.
To address this, we propose a novel compact layer called \modelfull (\model),
which combines local propagations and global convolutions.
The convolution filters in \model encompass all nodes, serving as a substitute for dense attention
in the graph domain. \model consists of four main components:
(1)~local propagation to capture local context,
(2)~global convolution to capture global context with quasilinear complexity,
(3)~data-controlled gating for context-specific operations on each node, and
(4)~positional/structural encoder for feature encoding and graph ordering.

Our evaluation has two main objectives:
\textbf{(O1):} Matching SOTA GT quality on small graph datasets emphasized by the community.
\textbf{(O2):} Scaling to larger graphs where traditional attention mechanisms
are impractical due to computational constraints.
Extensive evaluations across diverse benchmarks of varying tasks
and sizes demonstrate that \model scales to larger datasets and consistently delivers strong quality,
often achieving SOTA or competitive results.
The main contributions of work include:
\begin{itemize}
    \item We developed \model, a compact layer consisting of local and global
    context blocks with quasilinear time complexity.
    Unlike prior work, \model is a refined layer without intermediate
    parameters and non-linearities between local and global blocks,
    applying skip connections to the layer as a whole.
    \item To our knowledge, \model is the first to employ a global convolution model
    to develop a scalable and effective alternative to self-attention based GTs.
    Notably, it improves computational efficiency while preserving prediction quality,
    and in most cases, it leads to improvements.
    \item We demonstrated that \model scales to large-scale graphs that are infeasible for existing GTs utilizing self-attention due to their intrinsic quadratic complexity. \model further enhances prediction accuracy by up to 4.5\% for large graph datasets.
    \item We demonstrated \model's ability to capture long-range dependencies in graphs,
    achieving SOTA on the majority of long-range graph benchmark and improving results by up to $4.3\%$.
\end{itemize}

\section{Background and Related Work}

\subsection{Graph Neural Networks (GNNs)}

A graph $G = (V, E)$ comprises a set of vertices $V$ and edges $E \subseteq V \times V$. 
$A \in \mathbb{R}^{N \times N}$ is the adjacency matrix and a weighted edge
$(u \rightarrow v) \subseteq E$ exists between source $u$ and target
$v$ if $A_{u,v} \neq 0$.
The node feature matrix $X^{(0)} \in \mathbb{R}^{N \times d^{(0)}}$
maps $v$ to a feature vector $x_v^{(0)} \in \mathbb{R}^{d^{(0)}}$. 
$\mathcal{N}(v) = \{u \mid (u \to v) \in E \}$ is the incoming neighbors of $v$.
GNNs adopt an Aggregate-Combine framework~\citep{hamilton2017inductive}
to compute layer-$l$ representation $h_v^{(l)}$ such that:
\begin{align}
    \label{eq:gnn}
h_v^{(l)} = \mathrm{Combine}^{(l)}\big( \alpha_v^{(l)}, h_v^{(l-1)} \big), \qquad
\alpha_v^{(l)} &= \mathrm{Aggregate}^{(l)}\big( \big\{ h_u^{(l-1)} : u \in \mathcal{N}(v) \big\} \big)
\end{align}
Additionally, a pooling function generates graph representation,
$h_G = \mathrm{Pool} \big ( \big \{ h_v^{(L)} | v \in V \big \} \big)$.

\noindent{\textbf{Challenges.}}
GNNs efficiently scale to large graphs with linear complexity, $\mathcal{O}(|V| + |E|)$,
but they struggle with capturing long-range dependencies,
often requiring many hops and nonlinearities for information to traverse distant nodes
~\citep{alon2020bottleneck, dwivedi2022long}.
GTs effectively resolve this via dense pairwise attention.

\subsection{Graph Transformers (GTs)}
\label{sec:gt}

Graph Transformers (GTs) generalize Transformers~\citep{vaswani2017attention} to graphs.
At the core of Transformer lies the multi-head self-attention (MHA)~\citep{vaswani2017attention}, which maps the input $H \in \mathbb{R}^{N \times d}$ to $\mathbb{R}^{N \times d}$ as:
\begin{align}
    \label{eq:single_head}
    Attn(H) = \mathrm{Softmax}\left(\frac{QK^T}{\sqrt{d}}\right), \qquad y = \mathrm{SelfAttention}(H) = Attn(H) V
\end{align}
here \textit{query}
($Q = HW_q$), \textit{key} ($K = HW_k$), and \textit{value} ($V = HW_v$) are linear projections of the input, $W_q, W_k, W_v \in \mathbb{R}^{d \times d}$.
The attention matrix $Attn(H)$ captures the pair-wise similarities of the input.

Based on their \emph{focus of attention}, we group GTs into three:
\textbf{Sparse GTs} use the adjacency matrix as an attention mask, 
allowing nodes to pay attention to \emph{their neighbors},
which facilitates weighted neighborhood aggregation~\citep{gat2018, xu2018how}.
\textbf{Layer GTs} use a GNN to generate hop-tokens,
followed by MHA on these tokens, where nodes pay attention to \emph{their layer embeddings}~\citep{chen2023nagphormer,2024vcrgraphormer}.
While both Sparse and Layer GTs use attention,
they still struggle with long-range dependencies as their attention is restricted to fixed number of hops.
Comparatively, \textbf{Dense GTs} use attention on fully connected graph,
enabling nodes to pay attention to \emph{all the other nodes} regardless of their distances.
Various works have incorporated positional encodings (PE) into GTs
to provide topological information to otherwise graph-unaware models~\citep{dwivedi2020generalization,kreuzer2021rethinking,ying2021do,chen2022structure,zhao2023are,ma2023graph}. See \cref{appendix:related_work_comp} for related work details.

\noindent{\textbf{Challenges.}}
Dense GTs introduce computational and memory bottlenecks
due to their increased complexity from $\mathcal{O}(|V| + |E|)$ to $\mathcal{O}(|V|^2)$, restricting their application to large graphs.
GraphGPS~\citep{rampasek2022recipe} offers a modular framework that combines GNNs with a global attention module,
including subquadratic Transformer approximations~\citep{zaheer2020big, kreuzer2021rethinking}.
Unfortunately, the subquadratic models compromise quality while
MHA based ones struggle with scalability. 
Therefore,
finding a subquadratic attention replacement with a good quality remains a challenge,
and our work is dedicated to tackle this problem.

\noindent{\textbf{Other Related Work.}}
Exphormer~\citep{shirzad2023exphormer} enhances GraphGPS by using attention on expander graphs.
\citep{he2023generalization} generalizes ViT~\citep{dosovitskiy2020image} and MLP-Mixer~\citep{tolstikhin2021mlp}
to graphs.
\citep{zhang2022hierarchical} formulates an adversary bandit problem to sample nodes. %
HSGT~\citep{zhu2023hierarchical} learns multi-level hierarchies via coarsening.
GOAT~\citep{pmlr-v202-kong23a} uses dimensionality reduction to reduce computational cost of MHA.
\citep{diao2023relational} utilizes additional edge updates.

\subsection{Attention Alternatives}
Consider an input $u$ of length $N$ and a filter $z$. A circular convolution can be computed at each position $t$ of the input $u$, ranging from $0$ to $N-1$, as follows:
\begin{equation}
\label{eq:conv}
    y_t = (u * z)_t = \sum_{i=0}^{N-1} u_i z_{(t-i) \bmod N}
\end{equation}
where we assume a single-channel input and filter, which can be easily extended to multi-channel inputs.
CNNs~\citep{lecun1998gradient} optimize $z_t$ at every $K$ steps, where $K$ is a fixed filter size.
This \emph{explicit} parametrization captures local patterns within every $K$ steps. Alternatively, \emph{implicit} parameterizations represent the filter as a learnable function~\citep{tay2021long,gu2021efficiently,romero2022ckconv, smith2023simplified,fu2023simple,fu2023hungry}.
Convolutions can be efficiently computed through Fast Fourier Transform (FFT) in quasilinear time, offering a significant advantage.

A convolution is referred to as a global convolution when the filter has the same length as the input.
Global convolutional models have demonstrated the ability to capture longer contexts through pairwise interactions
(dot products) at any input position by proper filter parametrization, offering a promising alternative to attention~\citep{gu2021efficiently,romero2022ckconv,li2023what}.
Recently, Hyena~\citep{poli2023hyena} proposed a sequence model that combines short explicit convolutions and global implicit convolutions using a similar global filter design as CKConv and SGConv~\citep{romero2022ckconv, li2023what}, and it stands out by matching Transformer's quality in quasilinear time. Please refer to \cref{appendix:hyena} for details.
However, to the best of our knowledge, there has been no prior work dedicated to designing and utilizing global convolutional models for graphs.
\begin{figure*}
    \centering
    \includegraphics[width=\textwidth,trim=0cm 0cm 0cm 0cm]{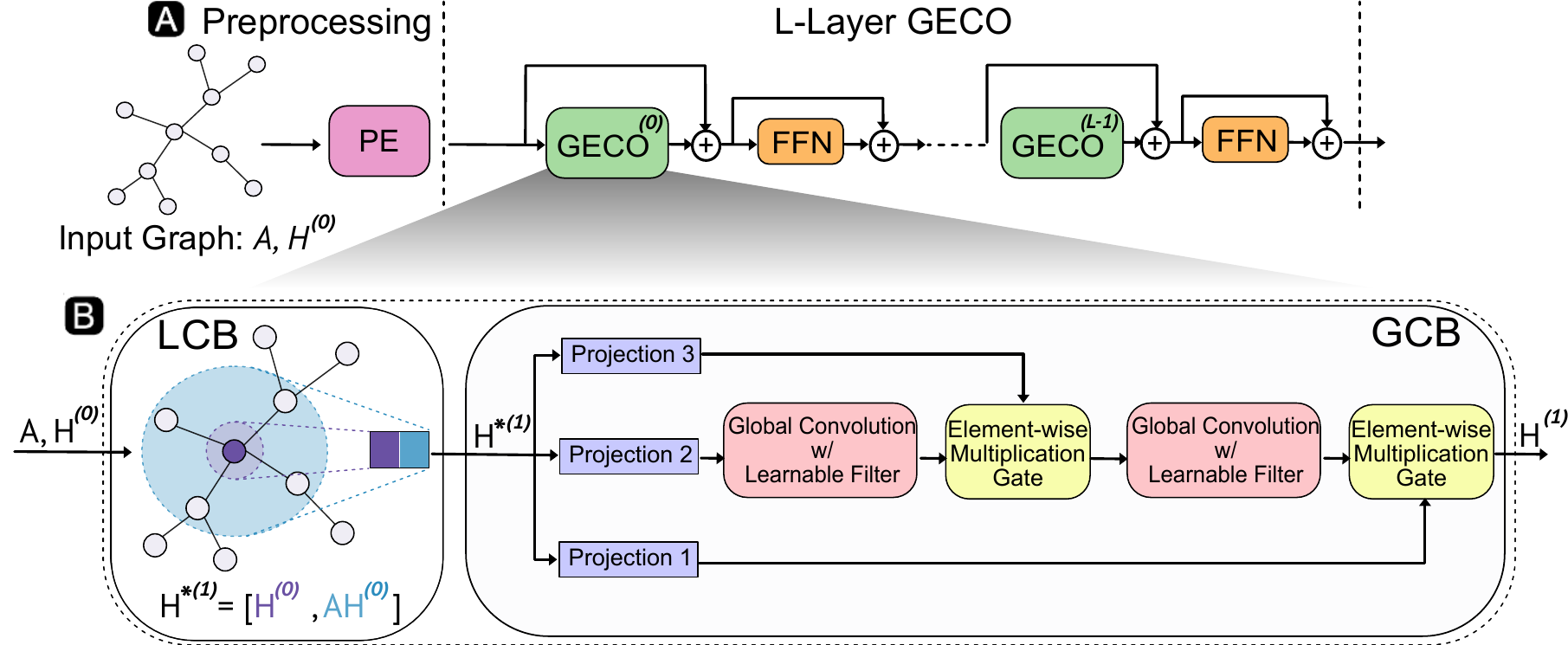}
    \caption{
        \inlinefig[11]{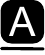} Our architecture comprises Positional Encoding (PE) block and
        Graph-Enhanced Contextual Operators ({\model}s) layers.
        PE adds positional encodings as a preprocessing step and each \model is followed by an FFN.
        \inlinefig[11]{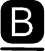}~A~\model layer contains a \lcb (\lcba) aggregating neighborhood embeddings and concatenating with originals to capture local dependencies, and a \gcb (\gcba) efficiently capturing global dependencies via global convolutions.
    }
    \label{fig:model_fig}
    \vspace{-1em}
\end{figure*}

\section{Proposed Architecture: \model}
\label{section:geco}

We present \modelfull (\model), a novel compact layer developed to replace dense attention with quasilinear time and memory complexity.
It draws inspiration from recent advancements in global convolutional models
and offers a promising approach to capture local and global dependencies with subquadratic operators.
Unlike Hyena, which focuses on sequences, \model is designed for graphs,
combining local propagations with global convolutions.
By utilizing the topological information of the adjacency matrix, it effectively captures local dependencies.
Furthermore, it introduces a new global convolution filter design for graphs to capture global dependencies.

As illustrated in Figure~\ref{fig:model_fig}, \model starts
with positional/structural encodings and proceeds through multiple layers of
\model, each followed by a feed-forward neural network (FFN).
We introduce the main components in the following subsections.

\subsection{Graph Structural/Positional Encodings}
\label{sec:positional_encoding}

Structural and positional encodings play a pivotal role in the realm of graph transformers.
In our approach, we follow the foundational work established in prior literature
concerning these encodings~\citep{dwivedi2020generalization, kreuzer2021rethinking, ying2021do, dwivedi2022graph}.
To seamlessly integrate these encodings with the original input features, we employ a concatenation method.
Given a positional/structural encoding matrix $U \in \mathrm{R}^{N \times d_u}$, where $d_u$ represents the encoding dimension,
we combine it with the original node features denoted by $X$.
This concatenation process results in a new feature matrix $X^*$, defined as follows: $X^* = [X, U]$.
For further details on incorporating relative encodings, please refer to Appendix~\ref{appendix:relative_encodings}.

\subsection{\lcb (\lcba)}
\lcb (\lcba) aggregates neighborhood embeddings for each node
and concatenates them with the original ones by utilizing the explicit topological information present in the adjacency matrix.
Notably, \emph{no parameters} are involved at this stage, making it akin to the traditional feature propagation
with a dense skip connection.
\lcba can be expressed as follows:
\begin{equation}
    h_v^{*(l)} = [h_v^{(l - 1)}, \alpha_v^{(l)}] \quad \textit{or} \quad H^{*(l)} = [H^{(l-1)}, AH^{(l-1)}]
\end{equation}
where $\alpha_v^{(l)}$ and $h_v^{(l - 1)}$ are defined as before.
Instead of adding self-edges for each node, we concatenate $\alpha_v^{(l)}$ and $h_v^{(l-1)}$,
enabling our model to distinguish node and propagation embeddings.
Moreover, rather than solely relying on $h_v^{(l)}$,
local attention mechanisms similar to those found in GAT~\citep{gat2018} can be incorporated. 
Alternative approaches for \lcba are further discussed in Section~\ref{sec:ablation}.

\begin{proposition} \label{prop:pb}
    \lcba can be computed in $\mathcal{O}(N + M)$ using Sparse Matrix Matrix (SpMM) multiplication between $X^{(l)}$ and $A$ in linear time complexity, where $M = |E|$.
\end{proposition}

\subsection{\gcb (\gcba)}
Efforts have aimed at creating efficient attention alternatives to capture longer contexts
via low-rank approximation, factorization, and sparsification,
often leading to trade-offs between efficiency and quality~\citep{10.1145/3564269}.
Meanwhile, recent sequence models
opt for linear convolutions or RNNs which offer near-linear time complexity~\citep{gu2021efficiently, fu2023hungry, peng2023rwkv, li2023what, poli2023hyena,nguyen2023hyenadna}.
Building upon the evolving research, for the first time, we explore whether global convolutions can capture global context within graph structures.

However, designed for sequences, many of the global convolutional models lack graph handling capabilities inherently.
This leads us to a key question: Can we develop an operator that effectively processes graphs using global convolutions?
Our investigation has yielded positive results, leading to \gcb (\gcba), a novel operator with graph awareness.
Below, we highlight the key distinctions and enhancements of \gcba compared to Hyena. Furthermore, we provide ablation studies and empirical comparisons in~\cref{sec:ablation} that demonstrate significant quality improvements.

\noindent \textbf{\emph{Graph-to-sequence:}}
Since we focus on graphs, we arrange both $A$ and $X$ using permutation $\pi$ and convert them into time-correlated sequences, aligning node IDs with time $(t)$.

\noindent \textbf{\emph{All-to-all information flow:}}
As our setup lacks causality unlike sequences,
we remove the causal mask from the global convolution kernel.
This allows information to flow mutually between all nodes, respecting the natural dynamics of graph data.
The non-causal filters are vital because the relationship between nodes is not inherently sequential or unidirectional.
Nodes can have mutual or bidirectional influences, and their relationships are not bound by a linear sequence like words in a sentence.

\noindent \textbf{\emph{Graph-aware context:}}
(1) The original proposal by~\citep{li2023what} for global convolutional models for sequences
involves exponential decay modulation for convolution filters, assigning higher weights to nearby points in the sequence.
In contrast, we aim to minimize the impact of the permutation $\pi$ during model training.
Therefore, we treat all nodes equally regardless of their distance under $\pi$ by eliminating this decay.
(2) Unlike Hyena~\citep{poli2023hyena}, \model does not employ short convolution along the sequence length,
as $\pi$ may not reflect a locality-sensitive order.
Instead, \model utilizes \lcba for local dependencies by leveraging the adjacency matrix.
In addition, we first apply \lcba before generating input projections, which further reduces the number of parameters in comparison to the prior work.

\noindent \textbf{\emph{Window of the global convolution:}}
We set the window size for global convolutions to match the number of nodes,
ensuring the inclusion of all nodes within the convolution operation.
Without the adjacency matrix, no explicit context is present for graphs.
Thus, shorter window lengths hold no meaningful interpretation.
This is similarly reasoned by the natural dynamics of graph data where node permutations
do not introduce proximity-based context.

\begin{wrapfigure}{R}{0.58\textwidth}
    \vspace{-20pt}
    \begin{minipage}[t]{0.58\textwidth}
    \begin{algorithm}[H]
    \small
    \caption{Forward pass of \gcba Operator}
    \label{alg:hyena_alt}
    \textbf{Input:} 
    \parbox{\linewidth}{Node embeddings $\X \in \mathbb{R}^{N \times d}$; 
    Order $K$;
    PE dim $d_e$;}
    \vspace{-1em}
    \begin{algorithmic}
    \STATE {\small 1.} $P_1, \dots, P_K, V = {\sf Projection}(\X)$ \textit{\# Linear projections $P_i$}
    \STATE {\small 2.} $F_1, \dots, F_K = {\sf Filter}(N, d_e)$ \textit{\# Position based filters $F_i$}\\
    \textit{\# Update $V$ until all projections are exhausted}
    \FOR{$i = 1,\dots, K$}
    \STATE {\small 3.} In parallel across $d$: $V_{t} \leftarrow (P_i)_{t}\cdot {\sf FFTConv}(F_i, V)_{t}$
    \ENDFOR  
    \STATE {\small 4.} Return $V$
    \end{algorithmic}
    \end{algorithm}
    \end{minipage}
    \vspace{-5pt}
\end{wrapfigure}
\cref{alg:hyena_alt} presents the \gcba (notations unified with~\citep{poli2023hyena}).
Given a node embedding matrix $X$,
\gcba generates $(K + 1)$ projections, where $K$ is a hyperparameter controlling its recurrence.
In this work, we set $K=2$, and in this case,
the three projections serve roles similar to query, key, and value.
For each projection, a filter is learned by a simple FFN, with node IDs used for filters' positional encoding.
Subsequently, the value $V$ is updated using global convolutions with one projection and filter at a time,
followed by element-wise multiplication gating, until all projections are processed.
\gcba is formally expressed as:
\begin{equation}
 \label{eq:gcba}
    y = v \odot (f_q * (q \odot (f_k * k))
\end{equation}
We assume single-channel features and omit layer notations for simplicity.
$q, k, v \in \mathbb{R}^{N \times 1}$ are linear projections of the input,
and $f_k, f_q \in \mathbb{R}^{N \times 1}$ are learnable filters with circular symmetry.
$\odot$ denotes Hadamard product (element-wise multiplication), and $*$ denotes circular convolution. 

\subsection{Surrogate Attention Analysis}
One natural question that arises is why the \gcba is a meaningful replacement for GT's self-attention.
To answer this, we can
rewrite the attention matrix as $Attn(H)~=~ \mathrm{Softmax}\left(\frac{HW_Q(HW_K)^T}{\sqrt{d}}\right)$ and
interpret it as a normalized adjacency matrix, where the pairwise similarity scores are edge weights learned through the attention mechanism.
\gcba with its modified filter design also learns a surrogate attention matrix that can be interpreted as an adjacency matrix
that takes the global context into account.
However, it is computed efficiently without storing the entire dense matrix.
Consequently, this design allows us to scale to larger datasets using the same computing resources.
Please refer to \cref{app:decomp} for details on the surrogate attention matrix decomposition.
\begin{proposition} \label{prop:hb}
    \gcba computes a surrogate attention matrix in $\mathcal{O}(N \log N)$ by using Fast Fourier Transform (FFT) and element-wise multiplication.
\end{proposition}

\subsection{Pitfalls of Permutation Sensitivity and Mitigation Strategies}

\label{sec:pitfalls}
The \model has certain pitfalls in terms of
permutation sensitivity.
While typical GNNs use permutation invariant functions~\citep{kipf2017semisupervised,
hamilton2017inductive}, \gcba's short and global convolutions
are shift-invariant but not permutation invariant.
Importantly, a line of research focuses on order-sensitive
GNNs~\citep{murphy2018janossy,murphy2019relational,chen2020can,sato2021random,huang2022going,chatzianastasis2023graph}
for enhanced expressibility.
Notably GraphSAGE~\citep{hamilton2017inductive} and \citep{moore2017deep}
with LSTM have shown outperforming results.
However, while these models may improve
quality for a specific task, 
the model could potentially lose its generalizability.
By replacing short convolutions with \lcba, we make the local mixing
permutation invariant. However, global convolutions remain order-sensitive.
To mitigate \gcba's permutation sensitivity, we have explored different
random permutation strategies.

\noindent \textbf{Static Random:}
We randomly permute the graph once before training as a naive baseline
and compare the performance variations between different runs.
Surprisingly, we observed that the final results are not significantly
impacted by different orderings, which we elaborate in \cref{sec:ablation}.

\noindent \textbf{Dynamic Random:}
With $N!$ permutations sampled, a permutation-sensitive function
can recover a permutation-invariant function~\citep{murphy2018janossy}.
Formally, consider parametrized function $\harrow{f}$ with parameters $W$,
and permutation $\pi$.
The original target permutation invariant function $\dbar{f}$ can be recovered as:
\begin{equation}
    \dbar{f}(X; W) = \frac{1}{N!} \sum_{\pi \in \Pi_{N}} \harrow{f}(A_{\pi}, X_{\pi}; W)
\end{equation}
However, $N!$ is intractable for large graphs, so one option is to sample permutations during training.
Consequently, we sample a random permutation per epoch per layer during model training.
Similar strategies have been also used for positional encodings,
such as random sign flipping for Laplacian PE~\citep{dwivedi2023benchmarking}.
This helps model to see many different permutations during training,
potentially memorize permutation invariance and gain robustness to different permutations.
$\pi\mathrm{-SGD}$~\citep{murphy2018janossy} theoretically proved such strategies approximates
$\dbar{f}$
with decreasing variance as more permutations are sampled.
\begin{proposition} \label{prop:permutation}
    \model with dynamic random sampling strategy is an approximate solution to original target permutation invariant function. 
\end{proposition}
For detailed proof, refer to \cref{app:permutation}. 
While we observe robustness to different orderings, 
understanding and addressing this limitation is crucial
for broadening the applicability of \model.
\vspace{-.5em}
\subsection{End-to-End Training}
\begin{wrapfigure}{R}{0.55\textwidth}
    \vspace{-25pt}
    \begin{minipage}{0.55\textwidth}
    \begin{algorithm}[H]
    \footnotesize
    \caption{End-to-end \texttt{\model} Model Training}
    \parbox{\linewidth}{\textbf{Input:}
    Adj. matrix $\Adj \in \mathbb{R}^{N \times N}$;
    Node features $\X \in \mathbb{R}^{N \times d}$;}
    \parbox{\linewidth}{
    Edge features $\E \in \mathbb{R}^{M \times d_e}$;}
    \vspace{-1em}
    \begin{algorithmic}
    \STATE {\small 1.} $\X, \Adj = \posenc(\X, \Adj, \E)$
    \FOR {$\ell = 0,\dots,L-1 $}
    \STATE {\small 2.} $\Per = \samplephi()$
    \STATE {\small 3.} $\X^{(0)} = \permute(\Adj, \X, \Per)$
    \STATE {\small 4.} $\X^{(l + 1)} = \layernorm( \model(\X^{(l)}, \Adj) + \X^{(l)})$\\
    \STATE {\small 5.} $\X^{(l + 1)} = \layernorm( \ffn(\X^{(l + 1)}) + \X^{(l + 1)}$)\\
    \ENDFOR
    \STATE {\small 5.} Return $\X^{L} \in \mathbb{R}^{N \times D}$
    \end{algorithmic}
    \label{algo:model_alg}
    \end{algorithm}
    \end{minipage}
    \vspace{-10pt}
\end{wrapfigure}
Algorithm~\ref{algo:model_alg} presents the end-to-end training with dynamic permutation.
We start by positional encodings.
The training is further broken into two main blocks.
\lcba propagates neighborhood embeddings and applies normalization, which is followed by \gcba.
Each \model is followed by an \ffn, such that
$\mathrm{FFN}(X) = \sigma(XW_1)W_2$,
where $W_1, W_2 \in \mathbb{R}^{d \times d }$ are the linear layer weights.
Both the \model and \ffn use skip connections, normalization, and dropout.
Alternatively, line 2 can be moved out of the for loop to achieve static permutation strategy to order the nodes as a preprocessing step.
\model uses three quasilinear operators and can be computed in $\mathcal{O}(N \log N + M)$.
For the complete algorithm and complexity analysis,
refer to \cref{appendix:algorithm,appendix:geco_comp} respectively.

\subsection{Comparison with Prior Work}
\noindent \textbf{{Hybrid Approaches.}}
Prior works~\citep{wu2021representing, dwivedi2020generalization,lin2021mesh,min2022transformer}
straightforwardly combine off-the-shelf GNNs and Transformers as separate local and global modules.
In contrast, \model's \lcba and \gcba are not auxiliary modules but
integrated components of a new compact layer design,
refining the model by removing intermediate parameters and non-linearities.
This design uses skip connections for the entire layer rather than separate components. 
Please refer to \cref{appendix:survey_comparison,appendix:gps_comparison} for details.

\noindent \textbf{{NAgphormer's Hop2Token}} is a preprocessing step decoupled from model training,
where feature propagation iterations are performed to generate node tokens~\citep{chen2023nagphormer}.
Such decoupling methods separate training from feature propagation,
hindering model to learn complex relationships between consecutive layers~\cite{wu2019simplifying}.
In contrast, \lcba's feature propagation is coupled with
learnable parameters of \gcba.
LCB is not a preprocessing step but rather an integral pre-step to GCB during model training.
In \cref{appendix:nagphormer}, we further discuss NAgphormer's recovery as a specific JKNets~\citep{xu2018representation} instance.

\noindent \textbf{Graph-Mamba}~\citep{wang2024graph} has recently adapted Mamba~\citep{gu2023mamba} for graphs.
It focuses on node ordering strategies based on prioritization
while using off-the-shelf permutation-sensitive components.
In contrast, we refine the layer design,
introduce \lcba,
aim to mitigate permutation-sensitivity, and further incorporate random permutation strategies
for improved robustness.
Notably, our evaluation also targets large node prediction datasets, unlike Graph-Mamba's focus on small graph-level tasks.

\textbf{Orthogonal research}:
(1) \emph{Model-agnostic methods:} Universally applicable feature encoding and initialization/tuning methods~\citep{dwivedi2020generalization,kreuzer2021rethinking,ying2021do,mialon2021graphit,chen2022structure,zhao2023are,tonshoff2023did,ma2023graph}.
(2) \emph{Scaling methods}
such as GOAT~\citep{pmlr-v202-kong23a}, HSGT~\citep{zhu2023hierarchical}, and LargeGT~\citep{dwivedi2023graph} that leverage self-attention.
GECO offers an alternative kernel combinable with these methods.
\cref{sec:ablation} provides evidence on when this combination is beneficial based on input size.
While Graph-ViT/MLP-Mixer propose alternatives, they are limited to graph-level tasks and
require graph re-partitioning at every epoch, which can be costly for large node-level tasks.
In contrast, \model does not require partitioning
and is applicable to node-level tasks as well.

\section{Experiments}
\label{section:experiments}

\begin{table*}[th!]
    \vspace{-.5em}
    \caption{LRGB Eval.: the \first{first}, \second{second}, and \third{third} are highlighted.
    We reuse the results from~\citep{rampasek2022recipe,shirzad2023exphormer}.
    }
    \label{tab:results_lrgb}
    \fontsize{8pt}{8pt}\selectfont
    \setlength\tabcolsep{4pt}
    \centering
    \begin{tabular}{lccccc}
    \toprule
        \multirow{2}{*}{\textbf{Model}} & \textbf{PascalVOC-SP} & \textbf{COCO-SP} & \textbf{Peptides-func} & \textbf{Peptides-struct} & \textbf{PCQM-Contact} \\
        \cmidrule{2-6}
        & \textbf{F1 score $\uparrow$} & \textbf{F1 score $\uparrow$} & \textbf{AP $\uparrow$} & \textbf{MAE $\downarrow$} & \textbf{MRR $\uparrow$} \\
        \midrule
        GCN & 0.1268 $\pm$ 0.0060 & 0.0841 $\pm$ 0.0010 & 0.5930 $\pm$ 0.0023 & 0.3496 $\pm$ 0.0013 & 0.3234 $\pm$ 0.0006 \\
        GINE & 0.1265 $\pm$ 0.0076 & 0.1339 $\pm$ 0.0044 & 0.5498 $\pm$ 0.0079 & 0.3547 $\pm$ 0.0045 & 0.3180 $\pm$ 0.0027 \\
        GatedGCN & 0.2873 $\pm$ 0.0219 & {0.2641 $\pm$ 0.0045} & 0.5864 $\pm$ 0.0077 & 0.3420 $\pm$ 0.0013 & 0.3218 $\pm$ 0.0011 \\
        GatedGCN+RWSE & 0.2860 $\pm$ 0.0085 & 0.2574 $\pm$ 0.0034 & 0.6069 $\pm$ 0.0035 & 0.3357 $\pm$ 0.0006 & 0.3242 $\pm$ 0.0008 \\
        \midrule
        Transformer+LapPE & 0.2694 $\pm$ 0.0098 & 0.2618 $\pm$ 0.0031 & 0.6326 $\pm$ 0.0126 & 0.2529 $\pm$ 0.0016 & 0.3174 $\pm$ 0.0020 \\
        SAN+LapPE & 0.3230 $\pm$ 0.0039 & 0.2592 $\pm$ 0.0158 & 0.6384 $\pm$ 0.0121 & 0.2683 $\pm$ 0.0043 & \third{0.3350 $\pm$ 0.0003} \\
        SAN+RWSE & {0.3216 $\pm$ 0.0027} & 0.2434 $\pm$ 0.0156 & 0.6439 $\pm$ 0.0075 & {0.2545 $\pm$ 0.0012} & {0.3341 $\pm$ 0.0006} \\
        GPS w/ Transformer & \third{0.3748 $\pm$ 0.0109} & \second{0.3412 $\pm$ 0.0044} & \second{0.6535 $\pm$ 0.0041} & \third{0.2500 $\pm$ 0.0005} & 0.3337 $\pm$ 0.0006 \\
        Exphormer & \second{0.3975 $\pm$ 0.0037} &  \first{0.3455 $\pm$ 0.0009}  &  \third{0.6527 $\pm$ 0.0043} & \second{0.2481 $\pm$ 0.0007} &  \first{0.3637 $\pm$ 0.0020}\\
        \midrule
        \model (Ours) & \first{0.4210 $\pm$ 0.0080} & \third{0.3320 $\pm$ 0.0032} & \first{0.6975 $\pm$ 0.0025} & \first{0.2464 $\pm$ 0.0009} & \second{0.3526 $\pm$ 0.0016} \\
        \bottomrule
    \end{tabular}
    \vspace{-1.5em}
\end{table*}

\subsection{Objective 1: Prediction Quality}
\label{sec:pred_quality}
We assess the \model on ten benchmarks outlined in~\Cref{appendex:dataset_g1},
where each dataset contains many small graphs, with
an average number of nodes ranging from tens to five hundred.
Consequently, \textbf{scalability is not a significant concern} for these datasets as
the computational load is determined by the average number of nodes.
As evidence, even the most computation-intensive GTs, such as Graphormer or GraphGPS with MHA,
can be trained on these datasets using Nvidia-V100 (32GB) or Nvidia-A100 (40GB) GPUs~\cite{ying2021do,rampasek2022recipe}.
The experiments in this section aim to demonstrate GECO's competitive predictive quality compared to GT baselines,
as many of them encounter memory or time issues with larger graphs. 
Nevertheless, for these evaluations, we begin by creating a hybrid GNN+\model by replacing the attention module used in GraphGPS.
For dataset and hyperparameter details please refer to Appendix~\ref{appendix:dataset}
and Appendix~\ref{appendix:hyperparameters}, respectively.

\noindent{\textbf{Long Range Graph Benchmark (LRGB).}}
Table~\ref{tab:results_lrgb} presents our evaluation on the LRGB,
a collection of graph tasks designed to test a model's ability to capture long-range dependencies. 
The results show that \model outperforms baselines across most datasets, with improvements up-to $4.3\%$.
For the remaining datasets, it ranks among the top three, with quality within $1.3\%$ of the best baseline.
By capturing long-range dependencies effectively, \model surpasses the performance of MHA in most cases without compromising quality.
Notably, \model's F1 score on PascalVOC increased from $0.4053$ to $0.4210$
without positional encodings, resulting in enhanced quality with a simplified model.

\begin{table*}[h]
    \caption{
    OGBG Eval.: \first{the best} is highlighted.
    We reuse the results from~\citep{rampasek2022recipe}.
    }
    \label{tab:results_ogb}
    \centering
    \fontsize{8pt}{8pt}\selectfont
    \begin{tabular}{lccccc}\toprule
    \multirow{2}{*}{\textbf{Model}} &\textbf{ogbg-molhiv} &\textbf{ogbg-molpcba} &\textbf{ogbg-ppa} &\textbf{ogbg-code2} \\\cmidrule{2-5}
    &\textbf{AUROC $\uparrow$} &\textbf{Avg.~Precision $\uparrow$} &\textbf{Accuracy $\uparrow$} &\textbf{F1 score $\uparrow$} \\\midrule
        SAN & 0.7785 $\pm$ 0.2470 & 0.2765 $\pm$ 0.0042 & -- & -- \\
        GraphTrans (GCN-Virtual) & -- & 0.2761 $\pm$ 0.0029 & -- & {0.1830 $\pm$ 0.0024} \\
        K-Subtree SAT & -- & -- & 0.7522 $\pm$ 0.0056 & \first{0.1937 $\pm$ 0.0028} \\
        GPS & 0.7880 $\pm$ 0.0101 & {0.2907 $\pm$ 0.0028} & \first{0.8015 $\pm$ 0.0033} & {0.1894 $\pm$ 0.0024} \\
        \midrule
        \model & \first{0.7980 $\pm$ 0.0200} & \first{0.2961 $\pm$ 0.0008} & {0.7982 $\pm$ 0.0042} & {0.1915 $\pm$ 0.002} \\
    \bottomrule
    \end{tabular}
\end{table*}

\noindent{\textbf{Open Graph Benchmark (OGB).}}
Table~\ref{tab:results_ogb} presents the evaluation results on OGB Graph level tasks.
For clarity, we only compare GT methods, but full table can be found in \cref{app:full_ogbg}.
Similar to GraphGPS, we observed instances of overfitting.
Nevertheless, \model outperforms GraphGPS on the majority of the datasets, except for ppa.
Across all datasets, it consistently secures the top two,
demonstrating its effectiveness as a high-quality and efficient GT alternative.

\begin{table*}[h]
    \caption{PCQM4Mv2 Eval.: the \first{first}, \second{second}, and \third{third} are highlighted.
    \emph{Validation} set is used for evaluation as \emph{test} set is private. We reuse results from~\citep{rampasek2022recipe}.
    }
    \label{tab:results_pcqm4m}
    \centering
    \fontsize{8pt}{8pt}\selectfont
    \setlength\tabcolsep{3.5pt}
    \begin{tabular}{lcccccccc|c} \toprule
    \textbf{PCQM4Mv2} & \textbf{GCN} & \textbf{GCN-virtual} & \textbf{GIN-virtual} & \textbf{GRPE} & \textbf{EGT} & \textbf{Graphormer} & \textbf{GPS-sm} & \textbf{GPS-med} &\textbf{\model} \\ \midrule
    \textbf{Train MAE $\downarrow$} & n/a & n/a & n/a & n/a & n/a & 0.0348 &  0.0653 &  0.0726 & 0.0578 \\ 
    \textbf{Val. MAE $\downarrow$} & 0.1379 & 0.1153  & 0.1083 & 0.0890 & 0.0869 & \third{0.0864} &  0.0938 & \second{0.0858} &\first{0.0841} \\ \midrule
    \textbf{\# Param.} & 2.0M & 4.9M  & 6.7M & 46.2M & 89.3M & 48.3M & 6.2M & 19.4M &  6.2M \\ 
    \bottomrule
    \end{tabular}
\end{table*}

\noindent{\textbf{PCQM4Mv2.}}
Table~\ref{tab:results_pcqm4m} 
demonstrates that \model outperforms both GNN and GT baselines on PCQM4Mv2 in terms of prediction quality.
Notably, \model uses only \textbf{1/8} and \textbf{1/3} of the parameters
required by Graphormer and GraphGPS, respectively.
This parameter reduction brings \model in close proximity to the parameter
count used by GNN baselines, while boosting their quality.
\vspace{-.5em}
\subsection{Objective 2: Scalability for Larger Graphs}
\label{sec:scalability_experiments}

\begin{table*}[h]
    \vspace{-.5em}
    \caption{Accuracy on large node prediction datasets: the \first{first}, \second{second}, and \third{third} are highlighted.
    We reuse the results from~\citep{han2023mlpinit, shirzad2023exphormer, zeng2021decoupling}, and run Exphormer locally except Arxiv.
    $-$ indicates that the data was either not included in the original work or could not be successfully reproduced.
    }
    \label{tab:result_large_datasets}
    \fontsize{8pt}{8pt}\selectfont
    \setlength\tabcolsep{4pt}
    \centering
    \begin{tabular}{lcccc}
    \toprule
         \multirow{2}{*}{\textbf{Model}} & \textbf{Flickr} & \textbf{Arxiv} & \textbf{Reddit} & \textbf{Yelp}  \\\cmidrule{2-5}
         \textbf{Accuracy} &\textbf{Accuracy} &\textbf{Accuracy} &\textbf{Accuracy}  &\textbf{Micro-F1 Score} \\\midrule
        GCN & 50.90 $\pm$ 0.12 & 70.25 $\pm$ 0.22 & 92.78 $\pm$ 0.11 & 40.08 $\pm$ 0.15  \\
        SAGE & \third{53.72 $\pm$ 0.16} & 72.00 $\pm$ 0.16 & \second{96.50 $\pm$ 0.03} & \third{63.03 $\pm$ 0.20}  \\
        GraphSaint & {51.37 $\pm$ 0.21} & 67.95 $\pm$ 0.24  & 95.58 $\pm$ 0.07 & 29.42 $\pm$ 1.32 \\
        Cluster-GCN & 49.95 $\pm$ 0.15 & 68.00 $\pm$ 0.59 & {95.70 $\pm$ 0.06} & {56.39 $\pm$ 0.64} \\
        GAT & 50.70 $\pm$ 0.32 & 71.59 $\pm$ 0.38 & \third{96.50 $\pm$ 0.11} & {61.58 $\pm$ 1.37} \\
        \midrule     
        Graphormer & OOM  & OOM & OOM &   OOM  \\
        Graphormer-SAMPLE  & 51.93 $\pm$ 0.21 &   70.43 $\pm$ 0.20  & 93.05 $\pm$ 0.22 5 &  60.01 $\pm$ 0.45 \\
        SAN  &  OOM & OOM & OOM & OOM  \\
        SAT  &  OOM & OOM & OOM & OOM  \\
        SAT-SAMPLE  & 50.48 $\pm$ 0.34 &    68.20 $\pm$ 0.46   &93.37 $\pm$  0.32 & {60.32 $\pm$ 0.65}  \\
        ANS-GT  & -- & 68.20 $\pm$ 0.46 & 95.30 $\pm$ 0.81 & -- \\
        GraphGPS w/ Transformer & OOM  & OOM & OOM & OOM  \\
        Exphormer & 52.60 $\pm$ 0.18 & \third{72.44 $\pm$ 0.28} & 95.90 $\pm$ 0.15 & 60.80 $\pm$ 1.56 \\
        HSGT  & \second{54.12 $\pm$ 0.51}        &  \second{72.58 $\pm$ 0.31}     &  --   &  \first{63.47 $\pm$ 0.45}         \\
        \midrule
        \model (Ours) & \first{55.55 $\pm$ 0.25} & \first{73.10 $\pm$ 0.24} & \first{96.65 $\pm$ 0.05} & \second{63.18 $\pm$ 0.59}   \\
        \bottomrule
    \end{tabular}
\end{table*}

We assess \model on $4$ benchmark datasets where each graph contains a much larger number of nodes.
Notably, traditional Dense GTs struggle to handle such large graphs due to their quadratic complexity
while GECO succeeds with its superior computational and memory efficiency.
In the following experiments, we design our models using only \model blocks, following
Algorithm~\ref{algo:model_alg}.
For simplicity, we avoid using structural/positional encodings as computing them may be infeasible for large graphs.
For details on datasets and hyperparameters, please refer to~\Cref{appendix:dataset_g2,appendix:hyperparameters}.

Unlike previous works that exhibit a trade-off between quality and scalability,
\model scales efficiently to large datasets and achieves superior quality across all compared to Dense GTs (Graphormer/GraphGPS), which suffer from OOM/timeout issues.
Remarkably, \model demonstrates significant predictive superiority, surpassing Dense GT baseline methods by up to 4.5\%.
On Arxiv, \model outperforms recently proposed GT works Exphormer and GOAT~\citep{pmlr-v202-kong23a} up to $0.7\%$.
Notably, Graphormer with sampling falls short in achieving competitive quality across all datasets.
When comparing \model to various baselines, including orthogonal methods, \model remains competitive.
It outperforms various baselines on Flickr, Arxiv, and Reddit, except for Yelp where the coarsening approach HSGT~\citep{zhu2023hierarchical} surpasses \model.
We leave the exploration of combining \model with orthogonal methods
such as expander graphs~\citep{shirzad2023exphormer}, hierarchical learning~\citep{zhu2023hierarchical},
and dimensionality reduction~\citep{pmlr-v202-kong23a} as future work to potentially get even better results.
Overall, the results highlight that the global context can enhance the modeling
quality for large node prediction datasets, justifying our motivation to find efficient high-quality attention alternatives.
To the best of our knowledge, \model is the first attempt to capture pairwise node relations without heuristics at scale.
Our evaluation illustrates its effectiveness as a Dense GT
alternative for large graphs.

\begin{table*}[h]
    \caption{Accuracy across large datasets with different permutation strategies (Natural/Static Random/Dynamic Random) with \model, alongside a comparison with default Hyena~\citep{poli2023hyena}.}
    \centering
    \label{tab:perm_ablation}
    \fontsize{8pt}{8pt}\selectfont
    \setlength\tabcolsep{4pt}
    \begin{tabular}{l|cccc}
    \toprule
    \textbf{Dataset} & \textbf{Hyena} & \textbf{\model} & \textbf{\model} & \textbf{\model}\\
    \midrule
    \textbf{Permutation} & \textbf{Natural} & \textbf{Natural} & \textbf{Static Random} & \textbf{Dynamic Random} \\
    \midrule
    Flickr & $46.97 \pm 0.08$ & $55.55 \pm 0.25$ & $55.73 \pm 0.27$ & $55.80 \pm 0.38$\\
    Arxiv & $56.04 \pm 0.61$ & $73.10 \pm 0.24$ & $73.08 \pm 0.28$ & $73.12 \pm 0.22$\\
    Reddit & $69.24 \pm 0.54$ & $96.65 \pm 0.05$ & $96.62 \pm 0.05$ & $96.68 \pm 0.06$\\
    Yelp & $50.08 \pm 0.31$ & $63.18 \pm 0.59$ & $63.23 \pm 0.50$ & $63.20 \pm 0.42$\\
    \bottomrule
    \end{tabular}
    \vspace{-2em}
\end{table*}
\subsection{Ablation Studies}
\label{sec:ablation}

\noindent{\textbf{Permutation Robustness.}}
In ~\cref{tab:perm_ablation}, we investigate \model's robustness to different permutations.
First, we maintained the natural ordering
of the graph and reported the mean and std of 10 runs with distinct seeds on this
fixed permutation (Natural).
Then, we repeated the same process,
but we have applied static and dynamic permutation strategies detailed in \cref{sec:pitfalls}.
The results indicate negligible differences between different strategies with
dynamic random showing slightly higher mean on multiple datasets.
However, all strategies seems to fall into similar confidence intervals,
hence we favor the simpler strategy in our experiments.

\noindent{\textbf{Hyena Comparison.}}
\cref{tab:perm_ablation} compares \model with the off-the-shelf Hyena by setting its filter size as the entire graph.
\model consistently outperforms the off-the-shelf Hyena with a significant margin.
This underscores the effectiveness of \model, particularly in its application of global convolutions for graph structures, and distinctly sets its apart from the off-the-shelf Hyena.

\begin{wraptable}{r}{0.55\textwidth}
    \vspace{-3em}
    \begin{minipage}{0.55\textwidth}
    \begin{table}[H]
        \caption{Ablation study on the LCB alternatives: the \first{first}, \second{second}, and \third{third} are highlighted. Conv-$x$ indicates 1D Convolution with a filter size of $x$.}
        \centering
        \fontsize{8pt}{8pt}\selectfont
        \setlength\tabcolsep{4pt}
        \label{tab:ablation_lrgb}
        \begin{tabular}{ll|ccc}
        \toprule
        \multirow{2}{*}{\makecell{\textbf{Model}}} &  \multirow{2}{*}{\makecell{\textbf{Local} \\ \textbf{Block}}} &\textbf{Pas.VOC-SP} &\textbf{Pep.-func} &\textbf{Pep.-struct} \\
        \cmidrule{3-5}
        & & \textbf{F1 $\uparrow$} &\textbf{AP $\uparrow$} &\textbf{MAE $\downarrow$} \\
        \midrule
        Transformer & N/A & \second{0.2762} & 0.6333 & \third{0.2525} \\
        Performer & N/A & 0.2690  & 0.5881 & 0.2739 \\
        \midrule
        \model & Conv-1  & \third{0.2752} & 0.6589 & 0.2587 \\
        \model & Conv-10 & 0.1757  & \second{0.6819} & \second{0.2516} \\
        \model & Conv-20 & 0.1645  & \third{0.6706} & 0.2534 \\
        \model & Conv-40 & 0.1445  & 0.6517 & 0.2547 \\
        \midrule
        \model & LCB & \first{0.3220}  &\first{0.6876} & \first{0.2454} \\
        \bottomrule
        \end{tabular}
    \end{table}
    \end{minipage}
    \vspace{-1em}
\end{wraptable}
\noindent{\textbf{\lcb Alternatives.}}
In \model, we adopted \lcba for graph-aware local context modeling instead of using 1D convolutions originally used in Hyena.
This is motivated by the limitation of 1D convolutions in capturing local dependencies in graphs where node order does not imply proximity.
At Table~\ref{tab:ablation_lrgb}, we focus on exploring alternatives to \lcba within our GECO module.
We experimented with replacing \lcba with 1D convolutions of various filter sizes to help understand its effectiveness.
We consistently observed a diminishing trend in
quality as filter sizes increased, which can be attributed to larger filter sizes leading to a mix of unrelated nodes within the graph.
In contrast, \model with \lcba consistently outperformed its alternatives as well as
the Transformer and Performer, highlighting its effectiveness in capturing local graph dependencies.

\begin{wrapfigure}{r}{0.45\textwidth}
    \centering
    \includegraphics[width=\linewidth]{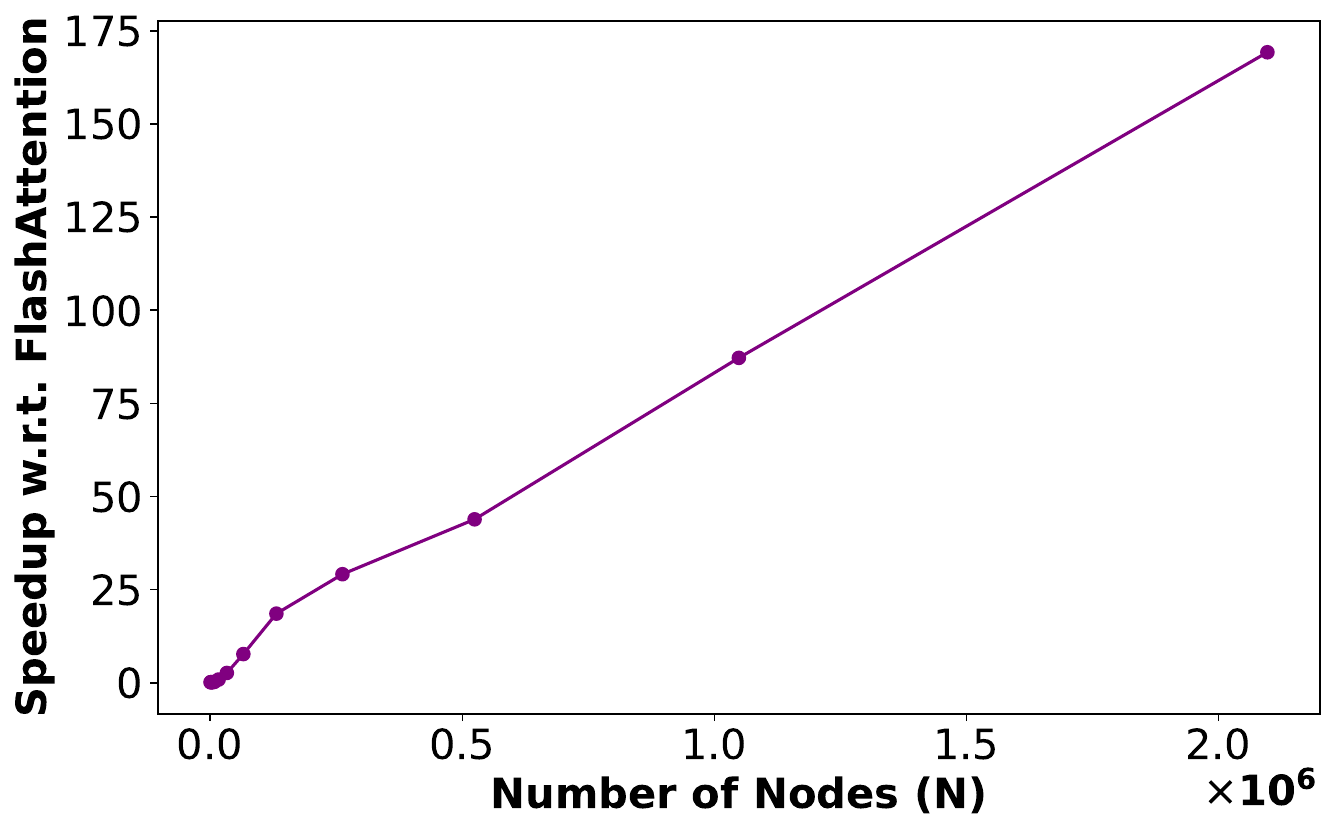}
    \caption{\small Relative speedup of GECO w.r.t. FlashAttention~\citep{dao2022flashattention} characterized by
    $\mathcal{O}(N/\log N)$}
    \label{fig:relative_speedup}
    \vspace{-1em}
\end{wrapfigure}
\noindent{\textbf{Scaling Study.}}
\cref{fig:relative_speedup} shows \model's speedup w.r.t. the optimized attention, FlashAttention~\citep{dao2022flashattention}, for increasing numbers of nodes using synthetic datasets
with similar sparsity patterns to those in \cref{tab:result_large_datasets}.
The results highlight that the speedup linearly increases with the number of nodes,
and \model reaches $169\times$ speedup on a graph with 2M nodes,
confirming its relative scalability.
Details including runtime numbers can be found in \cref{appendix:runtime_study}.

\vspace{-1em}
\section{Conclusion}
\vspace{-.5em}
We presented \model, a novel graph learning model that replaces
the compute-intensive MHA in GTs with an efficient and high-quality operator.
With comprehensive evaluation, we demonstrated
\model effectively scales to large datasets without compromising quality, and even outperforms it.
Moving forward, we plan to explore alternatives for \gcba,
and combinations with orthogonal approaches such as hierarchical learning.

\section*{Acknowledgement}
This work was partially done when the first author was a research scientist intern at Meta AI.
This work was partially supported by the NSF grant CCF-1919021.

\bibliography{geco_arxiv}

\newpage
\appendix
\onecolumn
\section{Datasets}
\label{appendix:dataset}

We gather a wide-ranging selection of $14$ datasets,
encompassing diverse graph types, tasks, and scales, collected from different sources.
To facilitate understanding, we classify these datasets into two overarching groups.
Each dataset in the first group comprises multiple graphs, each having small number of nodes/edges.
On the other hand, the second group comprises node prediction datasets,
each containing a single graph with much larger number of nodes.

\subsection{Datasets with Multiple Graphs}
\label{appendex:dataset_g1}

\begin{table}[h]
    \caption{Statics for the Datasets with Multiple Graphs, sorted by \#average nodes. \\\textbf{MRR}: Mean Reciprocal Rank, \textbf{AP}: Average Precision,
    \textbf{MAE}: Mean Absolute Error.\\Sparsity is calculated as $\frac{M}{N^2}$, where N and M represent the average number of nodes and edges, respectively.}
    \label{tab:dataset_g1}
    \fontsize{8pt}{8pt}\selectfont
    \setlength\tabcolsep{4pt}
    \begin{adjustwidth}{-2.5cm}{-2.5cm}
        \centering
    \begin{tabular}{lrrrrccccc}\toprule
    \multirow{2}{*}{\textbf{Dataset}} &\multirow{2}{*}{\textbf{\# Graphs}} &\multirow{2}{*}{\textbf{\# Avg. Nodes}} & \multirow{2}{*}{\textbf{\# Avg. Edges}} & \multirow{2}{*}{\textbf{Sparsity}} &\multirow{2}{*}{\textbf{Level}} &\multirow{2}{*}{\textbf{Task}} & \multirow{2}{*}{\textbf{Metric}} \\
    & & & & & &  &  \\
        \midrule
        \multicolumn{8}{l}{\textbf{Long Range Graph Benchmark}} \\
        \midrule
        PCQM-Contact  & $529,434 $  & $30.1$    & $61.0  $& \ensuremath{6.79 \times 10^{-2}}  & link & link ranking     & MRR \\
        Peptides-func & $15,535  $  & $150.9$   & $307.3 $& \ensuremath{1.36 \times 10^{-2}} & graph         & 10-task classif. & AP \\
        Peptides-struct & $15,535$  & $150.9$   & $307.3 $ & \ensuremath{1.36 \times 10^{-2}}   & graph         & 11-task regression & MAE \\
        COCO-SP       & $123,286 $  & $476.9$   & $2,693.7$& \ensuremath{1.20 \times 10^{-2}} & node& 81-class classif.& F1 \\
        PascalVOC-SP  & $11,355  $  & $479.4$   & $2,710.5$ & \ensuremath{1.20 \times 10^{-2}} & node& 21-class classif.& F1 \\
        \midrule
        \multicolumn{8}{l}{\textbf{Open Graph Benchmark}} \\
        \midrule
        PCQM4Mv2      & $3,746,620 $& $14.1 $   & $14.6$   & \ensuremath{7.25 \times 10^{-2}}  & graph         & regression       & MAE \\
        Molhiv   & $41,127 $   & $25.5  $  & $27.5$  & \ensuremath{4.29 \times 10^{-2}}    & graph         & binary classif.  & AUROC \\
        Molpcba  & $437,929$    & $ 26.0 $   & $28.1$  & \ensuremath{4.13 \times 10^{-2}}    & graph         & 128-task classif.& AP \\
        Code2    & $452,741 $  & $125.2  $ & $124.2$ & \ensuremath{1.59 \times 10^{-2}} & graph         & 5 token sequence & F1 \\
        PPA      & $158,100 $  & $243.4  $ & $2,266.1$ & \ensuremath{3.25 \times 10^{-2}} & graph         & 37-task classif. & Accuracy \\
    \bottomrule
    \end{tabular}
    \end{adjustwidth}
\end{table}

The first group consists of datasets used by GraphGPS~\citep{rampasek2022recipe}
and multiple other work in the community~\citep{he2023generalization, shirzad2023exphormer}.
This collection consists of datasets from
5 distinct sources,
and we further divide them into 2 groups: Long Range Graph Benchmark (LRGB)~\citep{dwivedi2022long} and Open Graph Benchmark (OGB)~\citep{hu2020ogb, hu2021ogb}.
For LRGB datasets, we respect the similar budget of ~500k parameters adopted by the previous literature~\citep{dwivedi2022long}.

\noindent{\textbf{Splits.}}
For these datasets, we employ the experimental setup used in GraphGPS for preprocessing and data splits,
please refer to the original work for details~\citep{rampasek2022recipe}.

\noindent{\textbf{Molhiv and Molpcba}}
are molecular property predictions sourced from
the OGB Graph (OGBG) collection~\citep{hu2020ogb}.
These predictions are derived from MoleculeNet, and each graph in the dataset
represents a molecule with atoms as nodes and chemical bonds as edges.
The primary task involves making binary predictions to determine whether
a molecule inhibits HIV virus replication or not.

\noindent{\textbf{Code2}} is a dataset collection consisting of Abstract Syntax
Trees (ASTs) extracted from Python definitions sourced from more than 13,000
repositories on GitHub. This dataset is also part of the OGBG.
The primary objective of this task is to predict the sub-tokens that compose the
method name. This prediction is based on the Python method body represented by the
AST and its associated node features.

\noindent{\textbf{PPA}} is a protein association network from OGBG.
It is compiled from protein-protein association networks originating from
a diverse set of 1,581 species, spanning 37 distinct taxonomic groups.
The primary objective of this task is to predict the taxonomic group from
which a given protein association neighborhood graph originates.

\noindent{\textbf{PascalVOC-SP and COCO-SP}}
are node classification datasets included in the LRGB collection.
These datasets are derived from super-pixel extraction on PascalVOC~\citep{everingham2010pascal} and MS COCO
datasets~\citep{lin2014microsoft} using the SLIC algorithm.
Each super-pixel (node) in these datasets is assigned to a specific class.

\emph{\textbf{PCQM4Mv2}} is a molecular (quantum chemistry) dataset obtained
from the OGB Large-Scale Challenge (LSC) collection, focusing on predicting the
DFT (density functional theory)-calculated HOMO-LUMO energy gap of molecules
using their 2D molecular graphs—a critical quantum chemical property~\citep{hu2021ogb}.
It is important to note that the true labels for the test-dev and test-challenge
dataset splits have been kept private by the challenge organizers to avoid any bias
in the evaluation process.
For our evaluation, we adopted the original validation set as our test set and
reserved a random sample of 150,000 molecules for the validation set,
following the experimental setting employed in GraphGPS.

\noindent{\textbf{Peptides-func, Peptides-struct, and PCQM-Contact}}
are molecular datasets from LRGB collection.
Peptides-func and Peptides-struct are both derived from $15,535$ peptides retrieved from SATPdb~\citep{singh2016satpdb},
but they differ in their task.
While Peptides-func is a graph classification task based on the peptide function,
Peptides-struct is a graph regression task based on the 3D structure of the peptides.
These graphs have relatively large diameters and are constructed in such a way that they necessitate
long-range interaction (LRI) reasoning to achieve robust performance in their respective tasks.
PCQM-Contact is derived from the PCQM4M~\citep{hu2021ogb}, which includes available 3D structures.
It has been filtered to retain only the molecules that were in contact at least once.
The main objective of this dataset is to identify, at the edge level, whether two molecules have been in contact.

\subsection{Datasets with Single Graph}
\label{appendix:dataset_g2}

\begin{table}[h]
    \caption{Overview of the graph learning dataset.}
    \label{tab:dataset_g2}
    \centering
    \footnotesize
    \begin{tabular}{lrrrrr}\toprule
   \textbf{Dataset} &\textbf{\# Nodes ($N$)} &\textbf{\# Edges ($M$)} &\textbf{Sparsity ($\frac{M}{N^2}$)} &\textbf{\# Features} &\textbf{\# Classes}\\\midrule
        Flickr        & $89,250$    & $899,756$  & $1.12 \times 10^{-4}$ & $500$ & $7$ \\
        ogbn-arxiv    & $169,343$	 & $1,166,243$ & $3.97 \times 10^{-5}$& $128$ &  $40$ \\
        Reddit         & $232,965$   & $114,615,892$ & $1.95 \times 10^{-6}$& $602$ & $41$ \\
        Yelp          & $716,847$   & $13,954,819$  & $2.26 \times 10^{-6}$& $300$ & $100$ \\
    \bottomrule
    \end{tabular}
\end{table}
The second group comprises large node classification datasets, and we utilize standard accuracy metrics for evaluation,
except for Yelp where we use micro-F1 following the general practice.
\noindent{\textbf{Splits.}} For the following datasets, we use publicly available standard splits across all datasets.

\noindent{\textbf{Reddit~\footnote{Reddit dataset is derived from the Pushshift.io Reddit dataset, which is a previously existing dataset extracted and obtained by a third party that contains preprocessed comments posted on the social network Reddit and hosted by pushshift.io.}}}~\citep{hamilton2017inductive} is a dataset derived from Reddit posts.
Each node in the dataset represents a post, and two posts are connected if they have been commented on by the same user.
The task is to classify which subreddit (community) a post belongs to.

\noindent{\textbf{Flickr}} and {\em \textbf{Yelp}} datasets are obtained from their respective networks~\citep{Zeng2020GraphSAINT}.
In the Flickr dataset, nodes represent uploaded images, and two nodes are connected if they share common properties or attributes.
In the Yelp dataset, two nodes are connected if they are considered friends within the social network.

\noindent{\textbf{OGBN- Arxiv}} is OGB Node Prediction (OGBN) dataset~\citep{hu2020ogb} which
is a citation network that connects Computer Science papers from Arxiv.
The features represent bag-of-word representations of the paper's title and abstract. The task is to identify the area of the papers.

\section{Architecture Details}

\subsection{Relative Encodings}
\label{appendix:relative_encodings}
In Section~\ref{sec:positional_encoding}, we discuss how to incorporate positional/structural encodings.
Importantly, our algorithm does not implicitly retain a dense attention matrix, making the integration of relative encodings more challenging.
We work with a relative encoding matrix $U_r \in \mathrm{R}^{N \times N}$, such as adjacency matrix or spatial information matrix,
and first create a low-rank approximation~\citep{jolliffe2016principal}
denoted by $U_r^* \in \mathrm{R}^{N \times d_r}$, where $d_r$ is the rank of the approximation.
Subsequently, we append the approximate relative encoding matrix to the node features and create an updated feature matrix $X^* = [X, U_r^*]$.
Note that, both node and edge positional/structural encodings can be extracted offline as a preprocessing step.

\subsection{End-to-end Training}
\label{appendix:algorithm}

\begin{algorithm}[h!]
\caption{{\sf Permute}}\label{alg:permute}
\textbf{Input:}
Adjacency matrix $\Adj \in \mathbb{R}^{N \times N}$; 
Node embeddings $\X \in \mathbb{R}^{N \times d}$; 
Node labels $\Y \in \mathbb{R}^{N \times d_y}$; 
Permutation $\Per \in \mathbb{R}^{N \times N}$;
\begin{algorithmic}
\STATE {\small 1.} $\Adj' \gets \Per \cdot \Adj \cdot \Per^\top$\\
\STATE {\small 2.} $\X' \gets \Per \cdot \X$\\
\STATE {\small 2.} $\Y' \gets \Per \cdot \Y$\\
\STATE Return $\Adj', \X', \Y'$
\end{algorithmic}
\end{algorithm}

Algorithm~\ref{alg:permute} performs a permutation operation on the graph
and node features based on the given permutation matrix and returns the permuted
node features, labels and adjaceny matrix.

    \begin{algorithm}[h!]
    \caption{{\sf \propagate}}\label{alg:propagate_block}
    \textbf{Input:}
    Adjacency matrix $\Adj \in \mathbb{R}^{N \times N}$; 
    Node embeddings $\X \in \mathbb{R}^{N \times d }$; 
    \begin{algorithmic}
    \STATE {\small 1.} $\hat{\Adj} = {\sf Normalized \; \Adj}$
    \STATE {\small 2.} $\X' = \hat{\Adj} \X$
    \STATE {\small 3.} Return $[\X, \X']$
    \end{algorithmic}
\end{algorithm}

The algorithm~\ref{alg:propagate_block} outlines the \lcb (\lcba),
which plays a key role in the process.
The \lcba starts by aggregating neighborhood embeddings, optionally a normalization can be applied.
Normalized $\Adj$ can be derived in different ways.
The standard GCN derives it as follows: $\hat{\Adj} = \D^{-1/2} \Adj \D^{-1/2}$,
where $\D^{-1/2}$ where $\D = \text{{diag}}(\Adj \mathbf{1})$, and $\mathbf{1}$ is a column vector of ones.
Preserving the original node features after propagation is essential.
While some models, like GCN, achieve this by introducing self-edges to the original graph,
this approach has a limitation: nodes treat their own embeddings and their neighbors'
embeddings equally in terms of importance.
To overcome this limitation, we adopt a different strategy.
We concatenate the original node embeddings with the propagated embeddings,
similar to a dense residual connection~\citep{huang2017densely}.
Notably, this step does not involve any learnable parameters.
In ~\cref{sec:ablation},
we explore several other variants with GNN models.

\begin{algorithm}[h!]
    \caption{{\sf Projection}}\label{alg:projection}
    \textbf{Input:}
    Node embeddings $\X \in \mathbb{R}^{N \times d }$; 
    \begin{algorithmic}
    \STATE {\small 1.} In parallel across $N$: $Z = {\sf Linear}(X)$, ${\sf Linear}:\R^{d} \rightarrow \R^{(K + 1)d}$ \\
    \STATE {\small 3.} Reshape and split $Z$ into $X_1, X_2, \dots, X_K, V$, where $X_k, V \in \R^{d \times N}$\\ 
    \STATE Return $X_1, X_2, \dots, X_K, V$
    \end{algorithmic}
\end{algorithm}

\begin{algorithm}[h!]
    \caption{${\sf \model}$ Operator}\label{alg:model_block}
    \caption{Forward pass of ${\sf \model}$}
    \textbf{Input:}
    Adjacency matrix $\Adj \in \mathbb{R}^{N \times N}$;
    Node embeddings $\X \in \mathbb{R}^{N \times d}$; 
    \begin{algorithmic}
    \STATE {\small 1.} $\X = \bn( \propagate(\X, \Adj) )$\\
    \STATE {\small 2.} $\X = GCB(\X, \Adj)$\\
    \STATE Return $\X$
    \end{algorithmic}
\end{algorithm}

\section{{Hyena Details}}
\label{appendix:hyena}
Equation~\ref{eq:conv} can be also expressed as $y_t = Tu$,
where $T\in \mathbb{R}^{N \times N}$ is Toeplitz kernel matrix induced by filter $z$.
Then the second-order Hyena operator with input $x \in \mathbb{R}^{N \times 1}$ defined as follows:
\begin{small}
\begin{align}
    \label{eq:hyena}
    y = \mathrm{Hyena}(x_1, x_2)x_3, \quad \mathrm{Hyena}(x_1, x_2) = D_{x_2} T D_{x_1}
\end{align}
\end{small}
where $x_1, x_2$, and $x_3$ are all projections of the input $x$,
and $T \in \mathbb{R}^{N \times N}$ is used as a learnable convolution filter.
In this context, $T$ is learned by a neural network, where
$T_{uv} = z_{u - v} = z_t = \gamma_\theta(t)$.
$D_{x_1}, {D_{x_2}} \in \mathrm{R}^{N \times N}$
are diagonal matrices with $x_1$ and $x_2$ on their diagonals respectively.

\noindent{\textbf{Connection to Attention.}}
$\mathrm{Hyena}(x_1, x_2)$ acts similar to attention matrix at Equation~\ref{eq:single_head},
however, it is realized by interleaving global convolutions and element-wise gating.
Furthermore, $y = \mathrm{Hyena}(x_1, x_2)x_3$ is efficiently computed without materializing the full matrix,
using FFT convolutions and gating.
Please refer to \citep{poli2023hyena} for more details.

\section{Proofs}
\subsection{Decomposition of the Surrogate Attention Matrix}
\label{app:decomp}
Let $q, k$ and $v$ be the linear projections of the input.
Moreover, for simplicity, assume $A$ is the surrogate attention matrix.
\citep{poli2023hyena} demonstrates that one can decompose the linear map $y = A^\psi_\varphi(q, k)v$ into a sequence of factors each dependent on a projection of the input such that  $A^\psi_\varphi(q, k)v = A^\psi(q)A_\varphi(k)$.
In this context, it is assumed that $D_q, D_k \in \mathrm{R}^{N \times N}$ are diagonal matrices with $q$ and $k$ are the respective entries in their main diagonal respectively.
Then, we can have the following:

\begin{equation}
\begin{split}
    A^\psi(q) &= D_q S_\psi\\
    A_\varphi(k) &= D_k S_\varphi\\
\end{split}
\end{equation}

Above, $S_\psi$ and $S_\varphi$ are Toeplitz matrices and are used as global convolution kernels with respective impulse responses. Moreover, the surrogate attention matrix is decomposed into two terms, $A^\psi(q)$ and $  A_\varphi(k)$ which are computed by multiplying the diagonal matrices with the Toeplitz matrices:

\begin{equation}
    \label{eq:decomp}
    A^\psi_\varphi(q, k)v = D_q S_\psi  D_k S_\varphi
\end{equation}

In the context of Hyena, the selection of $\psi$ and $\varphi$ matrices is chosen to be lower triangular, a choice well-suited for tasks involving causal language processing. In their publicly available repository~\footnote{https://github.com/HazyResearch/safari}, the authors also explore a bi-directional variant. However, in this particular usage scenario, the kernel sizes are doubled to ensure proper input padding. This results in the convolution kernel's length being twice that of the input, causing the weights to wrap around the input. This bi-directional convolution, while having bidirectional properties, is still fundamentally directional.

For our specific use case in graph applications, we configure the filter length to match the input length, corresponding to the number of nodes.
Furthermore, we ensure that the Toeplitz matrices are designed with circular symmetry, thereby ensuring a non-causal convolution. In simpler terms, this means that any two nodes can mutually influence each other, as information from the future (a node with a higher node ID) can affect the past (a node with a lower node ID). Consequently, we represent $S_\psi$ and $S_\varphi$ as matrices with circular Toeplitz symmetry:

\begin{equation}
    S_\psi = \begin{bmatrix}
        \psi_0 & \psi_{N-1} & \cdots & \psi_{2} & \psi_{1} \\
        \psi_1 & \psi_0 & \cdots & \vdots & \vdots \\
        \vdots & \ddots & \ddots & \psi_0 & \psi_{N-1} \\
        \psi_{N-1} & \psi_{N-2} & \cdots & \psi_1 & \psi_0
    \end{bmatrix}, \qquad
    S_\varphi = \begin{bmatrix}
        \varphi_0 & \varphi_{N-1} & \cdots & \varphi_{2} & \varphi_{1} \\
        \varphi_1 & \varphi_0 & \cdots & \vdots & \vdots \\
        \vdots & \ddots & \ddots & \varphi_0 & \varphi_{N-1} \\
        \varphi_{N-1} & \varphi_{N-2} & \cdots & \varphi_1 & \varphi_0
    \end{bmatrix}.
\end{equation}

Moreover, one can extend the \cref{eq:decomp} as the following using the symmetric filters:

\begin{equation}
    \label{eq:sur_att}
    \resizebox{\textwidth}{!}{
    $\begin{aligned}
        \underset{ 
        \begin{bmatrix}
            q_0 &  &  & &  \\
             & q_1 &  & & \\
             &  & \ddots & &\\
             &  &  & q_{N-2} &\\
             &  &  &  & q_{N-1} 
        \end{bmatrix}}{\displaystyle D_q}
        \underset{ 
        \begin{bmatrix}
        \psi_0 & \psi_{N-1} & \cdots & \psi_{2} & \psi_{1} \\
        \psi_1 & \psi_0 & \cdots & \vdots & \vdots \\
        \vdots & \ddots & \ddots & \psi_0 & \psi_{N-1} \\
        \psi_{N-1} & \psi_{N-2} & \cdots & \psi_1 & \psi_0
        \end{bmatrix}}{\displaystyle S_\psi}
        &
        \underset{ 
        \begin{bmatrix}
            k_0 &  &  & &  \\
             & k_1 &  & & \\
             &  & \ddots & &\\
             &  &  & k_{N-2} &\\
             &  &  &  & k_{N-1} 
        \end{bmatrix}}{\displaystyle D_k}
        \underset{
        \begin{bmatrix}
        \varphi_0 & \varphi_{N-1} & \cdots & \varphi_{2} & \varphi_{1} \\
        \varphi_1 & \varphi_0 & \cdots & \vdots & \vdots \\
        \vdots & \ddots & \ddots & \varphi_0 & \varphi_{N-1} \\
        \varphi_{N-1} & \varphi_{N-2} & \cdots & \varphi_1 & \varphi_0
        \end{bmatrix}}{\displaystyle S_\varphi} 
        \\
        \underset{\displaystyle A_\psi(q)}{ 
        = \begin{bmatrix}
            q_0 \psi_0 & q_0 \psi_{N-1}  & \cdots & q_0 \psi_{2}  &  q_0 \psi_{1}\\
            q_1 \psi_1 & q_1 \psi_0 & \cdots &\vdots & \vdots \\
            \vdots & \ddots & \ddots & q_{N-2} \psi_{0} & q_{N-2} \psi_{N-1} \\
            q_{N-1} \psi_{N-1} & q_{N-1} \psi_{N-2} & \cdots & q_{N-1} \psi_{1} & q_{N-1} \psi_0
        \end{bmatrix}}
        &
        \underset{\displaystyle A_\varphi(k)}{ 
        \begin{bmatrix}
            k_0 \varphi_0 & k_0 \varphi_{N-1}  & \cdots & k_0 \varphi_{2}  &  k_0 \varphi_{1}\\
            k_1 \varphi_1 & k_1 \varphi_0 & \cdots &\vdots & \vdots \\
            \vdots & \ddots & \ddots & k_{N-2} \varphi_{0} & k_{N-2} \varphi_{N-1} \\
            k_{N-1} \varphi_{N-1} & k_{N-1} \varphi_{N-2} & \cdots & k_{N-1} \varphi_{1} & k_{N-1} \varphi_0
        \end{bmatrix}}\\
    \end{aligned}$}
\end{equation}

Then, we can write the surrogate attention scores as:

\begin{equation}
    \label{eq:decomp_ij}
    A^\psi_\varphi(q, k)_{ij} = q_i \sum_{n=0}^{N-1} k_n \psi_{(i-n) \bmod N} \varphi_{(n-j) \bmod N}
\end{equation}

\subsection{Dynamic Random Permutation Sampling}
\label{app:permutation}
The Janossy pooling function $\dbar{f}$ is a framework for constructing permutation-invariant functions from permutation-sensitive ones, such as RNNs, CNNs, LSTMs.
In our case, we use it for global convolution models.
In the original work~\citep{murphy2018janossy}, it is formally defined as:
\begin{definition}[Janossy Pooling~\citep{murphy2018janossy}]
Consider a function $\harrow{f}: \sN \times \sH^\cup \times \sR^b \to \sF$ on variable-length but finite sequences $\vh$, parameterized by $\ftheta \in \sR^b$, $b > 0$. A permutation-invariant function $\dbar{f}: \mathbb{N} \times \sH^\cup \times \sR^b \to \sF$ is the Janossy function associated with $\harrow{f}$ if
\begin{equation}
    \label{eq:JanossyPoolingN}
\dbar{f}(|\vhid|, \vhid ; \ftheta ) = \frac{1}{|\vhid|!} \sum_{\pi \in \Pi_{|\vhid|}} \harrow{f}(|\vhid|, \vhid_{\pi}; \ftheta),
\end{equation}
where $\Pi_{|\vhid|}$ is the set of all permutations,
and $\vhid_\pi$ is a particular permutation of the sequence.
\end{definition}
The output of $\dbar{f}$ can further be chained with another neural network parameterized by $\theta^{(\rho)}$~\citep{murphy2018janossy}:

\begin{equation}
\label{eq:JanossyPoolingRho}
\dbar{y}(\vx ; \vtheta^{(\rho)}, \vtheta^{(f)}, \vtheta^{(h)}) = \rho\left(\frac{1}{|\vhid|!}
\sum_{\pi \in \Pi_{|\vhid|}} \harrow{f}(|\vhid|, \vhid_{\pi}; \ftheta); \vtheta^{(\rho)}\right), \text{where}  \quad \vhid \equiv h(\vx ; \vtheta^{(h)}).
\end{equation}

However, notice that $\Pi_{|\vhid|}$ becomes computationally intractable for large graphs.
To address this, \citep{murphy2018janossy} proposes the $\pi$-SGD,
a stochastic optimization procedure that approximates the original objective by randomly sampling input permutations during training.
Formally, $\pi$-SGD optimizes the modified objective $\dbar{\Jloss}$:
\begin{equation} \label{eq:RLoss}
    \begin{split}
    \dbar{\Jloss}(\train;\vtheta^{(\rho)}, \vtheta^{(f)}, \vtheta^{(h)})  &= \frac{1}{N}\sum_{i=1}^N E_{\rvs_i}\left[ 
     \! L\Bigg(\vy(i) , 
    \rho\Big( \harrow{f}(|\vhid^{(i)}|, \vhid^{(i)}_{\rvs_i}; \vtheta^{(f)}) ; \vtheta^{(\rho)} \Big) \Bigg)
    \right]\\
    &= \frac{1}{N}\sum_{i=1}^N  \frac{1}{|\vhid^{(i)}|!} \sum_{\pi \in \mathbb{P}i_{|\vhid^{(i)}|}} 
     L\Bigg(\vy(i) , \rho\Big( \harrow{f}(|\vhid^{(i)}|, \vhid^{(i)}_\pi; \vtheta^{(f)}) ; \vtheta^{(\rho)} \Big) \Bigg).
    \end{split}
\end{equation}
where at iteration $t$, the parameters $\vtheta_t$ are updated as:
\begin{equation}\label{eq:piSGD}
\vtheta_t = \vtheta_{t-1} - \eta_{t} \rmZ_t,
\end{equation}
where $\rmZ_t$ is the random gradient computed over a mini-batch $\gB$ by sampling random permutations $\rvs_i \sim \text{Uniform}(\Pi_{|\vhid^{(i)}|})$.

As demonstrated in \citep{murphy2018janossy}, $\dbar{\Jloss}$ is permutation-invariant.
If the function class modeling $\harrow{f}$ has sufficient expressive capacity to represent permutation-invariant functions,
then minimizing $\dbar{\Jloss}$ will also minimize the original objective $\dbar{L}$.
Furthermore, minimizing $\dbar{\Jloss}$ implicitly regularizes $\harrow{f}$ to learn permutation-insensitive functions.
\begin{proposition}[$\pi$-SGD Convergence]\label{p:piSGD}
Under similar conditions as standard stochastic gradient descent (SGD),
the $\pi$-SGD algorithm defined in Equation~\ref{eq:piSGD} enjoys properties of almost sure convergence to the optimal $\vtheta$ minimizing $\dbar{\Jloss}$ in Equation~\ref{eq:RLoss}, as proven in \citep{murphy2018janossy}.
\end{proposition}
Effectively, $\pi$-SGD is an instance of Robbins-Monro stochastic approximation of gradient descent that optimizes $\dbar{\Jloss}$ by sampling random permutations of the input data during model training~\citep{murphy2018janossy}.

Note that these proofs are based on the original work~\citep{murphy2018janossy}.
For further details and complete proofs, please refer to the original work.

\section{Computational Complexity Discussion}

\begin{table}[h]
    \centering
    \footnotesize
    \setlength\tabcolsep{4pt}
    \begin{tabular}{cccccc}
    \hline
    \textbf{Model} & \textbf{GNN} & \textbf{Dense GT} & \textbf{Layer GT} & \textbf{GraphGPS} & \textbf{\model} \\
    \hline
    \multirowcell{2}{Long-range \\ Modeling} & \multirow{2}{*}{$\times$} & \multirow{2}{*}{\checkmark}  & \multirow{2}{*}{$\times$}  & \multirow{2}{*}{\checkmark}  & \multirow{2}{*}{\checkmark}  \\
     & & & & & \\
    \hline
    Time & $\mathcal{O}(L(N + M))$ &  $\mathcal{O}(LN^2)$ &  $\mathcal{O}(L(N + M) + L^2)$ & $\mathcal{O}(LN^2)$ & $\mathcal{O}(L(N \log N + M))$\\
    \hline
    Memory & $\mathcal{O}(L(N + M))$ &  $\mathcal{O}(LN^2)$ & $\mathcal{O}(L(N + M))$ & $\mathcal{O}(LN^2)$ & $\mathcal{O}(L (N \log N + M))$\\
    \hline
    \end{tabular}
    \caption{Computational Complexity Comparison for Full Batch Training}    \label{table:model_comparison}
\end{table}

Table~\ref{table:model_comparison} presents an overview of the complexities associated with various models.
It is important to note that while we provide specific categorizations, certain models within those categories may exhibit different complexities.
Hence, the complexities presented here represent the general case.

\subsection{Related Work}
\label{appendix:related_work_comp}
\noindent{\textbf{GNN}} models~\citep{kipf2017semisupervised, hamilton2017inductive} can be evaluated efficiently
using sparse matrix operations in linear time with the number of nodes and edges in the graph.
Although, some other models such as Graph Attention Networks (GAT)~\citep{gat2018} and its variant GATv2~\citep{xu2018how}
which we categorize under \emph{Sparse GT} at Section~\ref{sec:gt} has higher complexity, due to the attention mechanism.
Asymptotically, these models can achieve higher computational efficiency compared to other methods such as
Dense GT and Layer GT. Furthermore, they encompass a rapidly expanding line of research
that investigates mini-batch sampling methods on GNNs~\citep{hamilton2017inductive, zou2019layer, Zeng2020GraphSAINT, zeng2021decoupling, Balin2023layerneighbor}.

\noindent{\textbf{Dense GT}}~\citep{dwivedi2020generalization, kreuzer2021rethinking, ying2021do, mialon2021graphit, chen2022structure}
involves pairwise attention between every node regardless of the connectivity of the graph,
and hence has quadratic complexity with the number of nodes $\mathcal{O}(N^2)$.
Recently, a variant of DenseGT called Relational Attention has been introduced,
which involves additional edge updates, further increasing the overall complexity to $\mathcal{O}(N^3)$.
Given that standard mini-batching methods cannot be applied to DenseGT, its application is limited to small datasets.

\noindent{\textbf{Layer GT.}} We categorize NAGphormer~\citep{chen2023nagphormer} under this category.
NAGphormer employs attention on the layer (hop) tokens and exhibits a complexity similar to GNNs,
given that the number of layers is typically fixed and smaller compared to the number of nodes or edges.
As discussed in detail in Section~\ref{appendix:nagphormer},
while NAGphormer may not completely address issues inherited in GNNs, it offers several highly desirable properties.
Firstly, it performs feature propagation as a preprocessing step, enabling the results to be reused.
Additionally, during training, NAGphormer does not require consideration of the connectivity,
which allows it to leverage off-the-shelf traditional mini-batch training methods,
thereby achieving parallelism and scalability on large datasets.

\noindent{\textbf{Hybrid GT}} includes recently proposed GraphGPS framework~\citep{rampasek2022recipe}.
GraphGPS combines the output of a GNN Module and Attention Module, and outputs of the two module is summed up within each layer.
We discuss the differences between \model and GraphGPS in detailed in Appendix~\ref{appendix:gps_comparison}.
The complexity of GraphGPS primarily depends on its attention module, which acts as a bottleneck.
Despite offering subquadratic alternatives for attention, the reported results indicate that the best performance
is consistently achieved when using the Transformer as the attention module.
Consequently, similar to Dense GT, GraphGPS exhibits a complexity of $\mathcal{O}(N^2)$, making it less suitable for large datasets.
In essence, GraphGPS serves as a versatile framework for combining GNN and attention, along with their respective alternatives.

\noindent{\textbf{Positional Encodings (PE)}}
play a crucial role for Graph Transformers.
GT~\citep{dwivedi2020generalization} incorporates Laplacian eigen-vectors (\lepe) as positional encodings (\pe) to
enrich node features with graph information. SAN~\citep{kreuzer2021rethinking} introduces learnable \lepe through
permutation-invariant aggregation. Graphormer~\citep{ying2021do} uses node degrees as \pe and 
shortest path distances as relative \pe, achieving remarkable success in molecular benchmarks.
GraphiT~\citep{mialon2021graphit} proposes relative \pe based on diffusion kernels. SAT~\citep{chen2022structure}
extracts substructures centered at nodes as additional tokens, while \citep{zhao2023are}
uses substructure-based local attention with substructure tokens. GOAT~\citep{pmlr-v202-kong23a} uses dimensionality reduction to reduce the computational cost of attention.
\citep{diao2023relational} enhances attention with additional edge updates.
While some PEs, like node degrees used in Graphormer~\citep{ying2021do},
can be efficiently computed, others, such as Laplacian eigen vectors, Laplacian PE, or all pairs shortest path (for relative PE),
involve computationally expensive operations, usually $\mathcal{O}(N^3)$ or higher.
The good news is step can be computed once as a preprocessed step and does not necessarily be computed in GPU.
Nevertheless, for extremely large graphs, computing PE can still be computationally infeasible.

\subsection{GECO}
\label{appendix:geco_comp}
    
\noindent{\textbf{\model}} is composed of two building blocks: \lcb (\lcba) and the \gcb (\gcba).
In Proposition~\ref{prop:pb}, we discuss the complexity of \lcba, which is $\mathcal{O}(N + M)$. This step exhibits a similar memory complexity of $\mathcal{O}(N + M)$.

\subsection{End-to-End Training}

Next, we analyze the complexity of \gcba in Proposition~\ref{prop:hb}, which is $\mathcal{O}(N \log N)$.
In the paper, we specifically focus on the case when the \gcba recurrence order is set to 2.
However, we can generalize this to $K$ recurrence, from which we derive the following complexity components:

\begin{enumerate}
    \item Each \gcba includes $(K + 1)$ linear projections, resulting in a complexity of $\mathcal{O}((K+1) N)$.
    \item Next, we have $K$ element-wise gating operations, contributing a complexity of $\mathcal{O}(K N)$.
    \item Finally, there are $K$ FFT convolutions, where both the input and filter sizes are $N$, resulting in a complexity of $\mathcal{O}(K N \log N)$.
\end{enumerate}

As a result, the generalized complexity of \gcba can be expressed as $\mathcal{O}(K N \log N)$.
Considering the end-to-end training complexity of \model, we can combine the complexities of \lcba and \gcba,
resulting in $\mathcal{O}(K N \log N + M)$.

Next, we examine the memory complexity of \model, with a particular focus on the FFT convolutions used.
The standard PyTorch~\citep{paszke2019pytorch} FFT Convolution typically requires $\mathcal{O}(N \log N)$ memory.
However, it is possible to optimize this complexity to $\mathcal{O}(N)$ by leveraging fused kernel implementations
of FFT Convolutions~\citep{fu2023hungry}.
As a result, we can express \model's memory complexity as $\mathcal{O}(KN + M)$ when utilizing these fused FFT Convolution implementations,
where $K$ is the recurrence order.

\section{Experimental Details}

\subsection{Baselines}
Our baselines through~\cref{tab:result_large_datasets,tab:results_ogb,tab:results_pcqm4m,tab:ablation_lrgb}
include GCN~\citep{kipf2017semisupervised}, Graphormer~\citep{ying2021do}, GIN~\citep{xu2018how}, GAT~\citep{gat2018}, GatedGCN~\citep{bresson2017residual,dwivedi2023benchmarking}, PNA~\citep{corso2020principal}, DGN~\citep{beaini2021directional},
CRaW1~\citep{toenshoff2021graph}, GIN-AK+~\citep{zhao2022from}, SAN~\citep{kreuzer2021rethinking}, SAT~\citep{chen2022structure},
EGT~\citep{hussain2022global}, GraphGPS~\citep{rampasek2022recipe}, and Exphormer~\citep{shirzad2023exphormer}.

Our baselines for~\cref{tab:result_large_datasets} include GCN~\citep{kipf2017semisupervised}, GraphSAGE~\citep{hamilton2017inductive},
GraphSaint~\citep{Zeng2020GraphSAINT}, ClusterGCN~\citep{chiang2019cluster}, GAT~\citep{gat2018}, JK-Net~\citep{xu2018representation}, GraphGPS~\citep{rampasek2022recipe} and Exphormer~\citep{shirzad2023exphormer},
Graphormer~\citep{ying2021do}, SAN~\citep{kreuzer2021rethinking}, SAT~\citep{chen2022structure},
ANS-GT~\citep{chien2021adaptive}, GraphGPS~\citep{rampasek2022recipe}, HSGT~\citep{zhu2023hierarchical}.

\subsection{Implementation and Compute Resources}

\noindent{\textbf{Implementation.}}
We have implemented our model using PyTorch-Geometric~\citep{Fey/Lenssen/2019},
GraphGPS~\citep{rampasek2022recipe} and Safari libraries~\citep{poli2023hyena}.
For the evaluation at~\cref{sec:pred_quality}, we have integrated our code into the GraphGPS framework as a global attention module.
For the evaluation at~\cref{sec:scalability_experiments}, we have implemented our own framework to efficiently run
large datasets~\footnote{Our implementations will be open-sourced during or after the double-blind review process}.

\noindent{\textbf{Compute Resources.}}
We have conducted our experiments on an NVIDIA DGX Station A100 system with 4 80GB GPUs.

\subsection{Hyperparameters}
\label{appendix:hyperparameters}

\noindent{\textbf{Baselines.}}
For the baseline results presented at~\cref{tab:result_large_datasets,tab:results_ogb,tab:results_pcqm4m,tab:ablation_lrgb}
we reuse the results from GraphGPS and Exphormer~\citep{rampasek2022recipe, shirzad2023exphormer}.
Moreover, for the baseline results presented at \cref{tab:result_large_datasets},
we present the previously reported results in the literature~\citep{han2023mlpinit, shirzad2023exphormer, zeng2021decoupling,zhu2023hierarchical}.

\noindent{\textbf{\model.}}
For datasets at~\cref{tab:dataset_g1}, we drop in and replace the global attention module with \model.
The missing results are marked by $-$.
Our choice of hyperparameters is guided by GraphGPS and Exphormer~\citep{rampasek2022recipe, shirzad2023exphormer}.
We started with the hyperparameters recommended by the related work including optimizer configurations,
positional encodings, and structural encodings.
Then we proceed to hand-tune some optimizer configurations, dropout rates, and hidden dimensions by simple line search
by taking validation results into account.
On multiple datasets including PascalVOC, COCO, molpcha, and code2, we have eliminated the positional and structural encodings.

For datasets at~\cref{tab:dataset_g2}, we have used hyperparameter optimization framework Optuna~\citep{optuna_2019}
with Tree-structured Parzen Estimator algorithm for hyperparameter suggestion with each tuning trial using a random seed.
We reported the test accuracy achieved with the best validation configuration over $10$ random seeds.
As part of our public code release, we will provide all configuration files detailing our hyperparameter choices. 

\section{Additional Experiments Discussions}
\label{appendix:additional_discussions}

\subsection{Extended Runtime Study}
\label{appendix:runtime_study}

\begin{table}[h!]
    \centering
    \footnotesize
    \caption{Runtime Comparison of GECO and FlashAttention~\citep{dao2022flashattention} on synthetic datasets.  $\mathcal{O}(N^2/(N \log N)) = \mathcal{O}(N/\log N)$ characterizes the speedup. Sparsity factor of each graph is set as $10/N$.}
    \begin{tabular}{rrrr}
        \toprule
        N &  GECO (ms) &  FlashAttention (ms) &  Relative Speedup \\
        \midrule
        $512$ &$       1.88$ &$                 0.27$ &$              0.14$ \\
        $1,024$ &$       2.13$ &$                 0.32$ &$              0.15$ \\
        $2,048$ &$       2.11$ &$                 0.31$ &$              0.15$ \\
        $4,096$ &$       2.42$ &$                 0.32$ &$              0.13$ \\
        $8,192$ &$       2.12$ &$                 0.51$ &$              0.24$ \\
        $16,384$ &$       2.13$ &$                 1.84$ &$              0.86$ \\
        $32,768$ &$       2.63$ &$                 6.92$ &$              2.63$ \\
        $65,536$ &$       3.73$ &$                28.74$ &$              7.70$ \\
        $131,072$ &$       6.21$ &$               115.23$ &$             18.56$ \\
        $262,144$ &$      15.74$ &$               458.64$ &$             29.14$ \\
        $524,288$ &$      41.72$ &$              1,830.29$ &$             43.87$ \\
        $1,048,576$ &$      83.90$ &$              7,317.04$ &$             87.21$ \\
        $2,097,152$ &$     173.15$ &$             29,305.77$ &$            169.25$ \\
        \bottomrule
    \end{tabular}
\end{table}

In this subsection, we provide details on runtime ablation study in \cref{sec:ablation}.

\noindent{\textbf{Experimental Setting.}} We have generated random graphs using Erd\H{o}s-R\'enyi
model. We increased the number of nodes from 512 to 4.2 million by doubling the number of nodes at
consecutive points, using a total of 14 synthetic datasets.
Furthermore, we set the sparsity factor of each graph to $10/N$, where N is the number of nodes as defined before,
aligning the sparsity of the graph with that of large node prediction datasets in Table \ref{tab:dataset_g2}.
Additionally, we fixed the number of features at 108 across all datasets.
We utilized publicly available FlashAttention implementation~\citep{dao2022flashattention}\footnote{https://github.com/Dao-AILab/flash-attention}.
For FlashAttention, we used 4 heads.
In both \model and FlashAttention, the number of hidden units is set as the number of features.

The results demonstrate that as the number of nodes grows larger, \model achieves significant speedups with respect to FlashAttention.
This is anticipated due to \model's complexity of $\mathcal{O}(N\log N + M)$,
while attention's complexity is $\mathcal{O}(N^2)$.
Considering the sparsity of real-world datasets, $N$ becomes a dominant factor, leading to a speedup characterized by $\mathcal{O}(N/\log N)$.

In summary, our findings support \model's efficiency for larger-scale applications, whereas for smaller scales,
the choice between the two could be influenced by factors beyond just performance.
As discussed throughout \cref{section:intro}-\cref{section:experiments},
the scalability gains are not expected for small graphs.
This is mostly due to low hardware utilization incurred by available FFT implementations.
However, \model still remains a promising approach due to
its high prediction quality, as we demonstrated in \cref{sec:pred_quality}.
On the other hand, on larger graphs, \model exhibits significant scalability, as demonstrated in \cref{sec:scalability_experiments}.
It consistently outperforms dense GTs on all large datasets and remains superior
or competitive when compared to the orthogonal approaches.

\subsection{PCQM4Mv2}
\label{app:full_PCQM}
\begin{table}[h!]
    \label{tab:full_PCQM}
    \caption{PCQM4Mv2 evaluation: the \first{first}, \second{second}, and \third{third} best are highlighted.
    \emph{Validation} set is used for evaluation as \emph{test} labels are private. We reuse results reported by~\citep{rampasek2022recipe}.
    }
    \centering
    \fontsize{8pt}{8pt}\selectfont
    \begin{tabular}{lccccc}\toprule
    \multirow{2}{*}{\textbf{Model}} &\multicolumn{3}{c}{\textbf{PCQM4Mv2}} \\\cmidrule{2-5}
    &\textbf{Test-dev MAE $\downarrow$} &\textbf{Validation MAE $\downarrow$} &\textbf{Training MAE $\downarrow$} &\textbf{\# Param.} \\\midrule
    GCN &0.1398 &0.1379 & n/a &2.0M \\
    GCN-virtual &0.1152 &0.1153 & n/a &4.9M \\
    GIN &0.1218 &0.1195 & n/a &3.8M \\
    GIN-virtual &0.1084 &0.1083 & n/a &6.7M \\\midrule
    GRPE &0.0898 &0.0890 & n/a &46.2M \\
    EGT &0.0872 &0.0869 & n/a &89.3M \\
    Graphormer &n/a &\third{0.0864} &0.0348 & 48.3M \\
    GPS-small &n/a &0.0938 &0.0653 &6.2M \\
    GPS-medium &n/a &\second{0.0858} &0.0726  &19.4M \\ \midrule
    \model &n/a &\first{0.08413} & 0.05782  &6.2M \\
    \bottomrule
    \end{tabular}
\end{table}

Due to space limitations, we have included only necessary information in \cref{sec:pred_quality} in \cref{tab:results_pcqm4m}. \cref{tab:full_PCQM} presents additional details.

\subsection{Open Graph Benchmark Graph Level Tasks}
\label{app:full_ogbg}
\begin{table*}[h]
    \label{tab:full_ogbg}
    \caption{
    OGBG Eval.: the \first{first}, \second{second}, and \third{third} are highlighted.
    We reuse the results from~\citep{rampasek2022recipe}.
    }
    \centering
    \fontsize{8pt}{8pt}\selectfont
    \begin{tabular}{lccccc}\toprule
    \multirow{2}{*}{\textbf{Model}} &\textbf{ogbg-molhiv} &\textbf{ogbg-molpcba} &\textbf{ogbg-ppa} &\textbf{ogbg-code2} \\\cmidrule{2-5}
    &\textbf{AUROC $\uparrow$} &\textbf{Avg.~Precision $\uparrow$} &\textbf{Accuracy $\uparrow$} &\textbf{F1 score $\uparrow$} \\\midrule
        GCN+virtual node & 0.7599 $\pm$ 0.0119 & 0.2424 $\pm$ 0.0034 & 0.6857 $\pm$ 0.0061 & 0.1595 $\pm$ 0.0018 \\
        GIN+virtual node & 0.7707 $\pm$ 0.0149 & 0.2703 $\pm$ 0.0023 & 0.7037 $\pm$ 0.0107 & 0.1581 $\pm$ 0.0026 \\
        GatedGCN-LSPE & -- & 0.2670 $\pm$ 0.0020 & -- & -- \\
        PNA & 0.7905 $\pm$ 0.0132 & 0.2838 $\pm$ 0.0035 & -- & 0.1570 $\pm$ 0.0032 \\
        DeeperGCN & 0.7858 $\pm$ 0.0117 & 0.2781 $\pm$ 0.0038 & {0.7712 $\pm$ 0.0071} & -- \\
        DGN & {0.7970 $\pm$ 0.0097} & 0.2885 $\pm$ 0.0030 & -- & -- \\
        GSN (directional) & \second{0.8039 $\pm$ 0.0090} & -- & -- & -- \\
        GSN (GIN+VN base) & 0.7799 $\pm$ 0.0100 & -- & -- & -- \\
        CIN & \first{0.8094 $\pm$ 0.0057} & -- & -- & -- \\
        GIN-AK+ & 0.7961 $\pm$ 0.0119 & \third{0.2930 $\pm$ 0.0044} & -- & -- \\
        CRaWl & -- & \first{0.2986 $\pm$ 0.0025} & -- & -- \\
        ExpC & 0.7799 $\pm$ 0.0082 & 0.2342 $\pm$ 0.0029 & \third{0.7976 $\pm$ 0.0072} & -- \\
        \midrule
        SAN & 0.7785 $\pm$ 0.2470 & 0.2765 $\pm$ 0.0042 & -- & -- \\
        GraphTrans (GCN-Virtual) & -- & 0.2761 $\pm$ 0.0029 & -- & {0.1830 $\pm$ 0.0024} \\
        K-Subtree SAT & -- & -- & 0.7522 $\pm$ 0.0056 & \first{0.1937 $\pm$ 0.0028} \\
        GPS & 0.7880 $\pm$ 0.0101 & {0.2907 $\pm$ 0.0028} & \first{0.8015 $\pm$ 0.0033} & \third{0.1894 $\pm$ 0.0024} \\
        \midrule
        \model & \third{0.7980 $\pm$ 0.0200} & \second{0.2961 $\pm$ 0.0008} & \second{0.7982 $\pm$ 0.0042} & \second{0.1915 $\pm$ 0.002} \\
    \bottomrule
    \end{tabular}
\end{table*}

As we wanted to focus on direct comparison between graph transformer models and due to space limitations, we have included only necessary information in \cref{sec:pred_quality} in \cref{tab:results_ogb}. \cref{tab:full_ogbg} presents additional details.
\model still consistently secures top three when other GNN-based models are included.
Some of these methods, such as CRaWl~\citep{toenshoff2021graph}, incorporate features like random walk features that can be used in other approaches as well.
In our comparison, we focus on evaluating \model's effectiveness with respect to attention or its alternatives.

\subsection{Comparison with various Graph Transformers Variants}
\label{appendix:survey_comparison}

\begin{table}[h!]
    \centering
    \caption{Comparison with various Graph Methods from~\citep{min2022transformer}.}
    \label{tab:gt_survey}
    \footnotesize
    \begin{tabular}{lc|cccc}\toprule
    \multirow{2}{*}{\textbf{Model}} & & {\textbf{molhiv}} & {\textbf{molpcba}} & \textbf{Flickr} & {\textbf{ogbn-arxiv}} \\\cmidrule{3-6}
    & & ROC-AUC↑ & AP↑  & Acc↑ &  Acc↑ \\\midrule
    TF & vanilla & 0.7466 & 0.1676 & 0.5279 & 0.5598 \\
    \midrule
    GA & before & 0.7339 & 0.2269 & 0.5369 & 0.5614 \\
    & alter & 0.7433 & 0.2474 & \third{0.5374} & 0.5599 \\
    & parallel & \third{0.7750} & 0.2444 & \second{0.5379} & 0.5647 \\
    \midrule
    PE & degree & 0.7506 & 0.1672 & 0.5291 & 0.5618 \\
    & eig & 0.7407 & 0.2194 & 0.5278 & \third{0.5658} \\
    & svd & 0.7350 & 0.1767 & 0.5317 & \second{0.5706} \\
    \midrule
    AT & SPB & 0.7589 & \third{0.2621} & 0.5368 & 0.5605 \\
    & PMA & 0.7314 & 0.2518 & 0.5288 & 0.5571 \\
    & Mask-1 & \second{0.7960} & \second{0.2662} & 0.5300 & 0.5598 \\
    & Mask-n & 0.7423 & 0.2619 & 0.5359 & 0.5603 \\
    \midrule
    GECO (Ours) & & \first{0.7980 $\pm$ 0.0200} & \first{0.2961 $\pm$ 0.0008} & \first{0.5555 $\pm$ 0.0025} & \first{0.7310 $\pm$ 0.0024}\\
    \bottomrule
    \end{tabular}
\end{table}

The survey~\citep{min2022transformer} classifies existing methods for enhancing
Transformer's awareness of topological structures into three main categories: 
1) Integrating GNNs as auxiliary modules (GA), (2) Enhancing positional embeddings from graphs (PE),
and (3) Improving attention matrices using graphs (AT).

Regarding GA, the survey explores different approaches to combining off-the-shelf GNNs with Transformer.
These methods involve adapting a series of GNNs and then applying a series of Transformers sequentially (before)~\citep{wu2021representing},
integrating GNNs and Transformers consecutively (Alternatively)~\citep{lin2021mesh}, or utilizing them in parallel at each layer~\citep{rampasek2022recipe}.
Notably, these models straightforwardly merge pre-existing GNN and Transformer models,
resulting in separate parameters and intermediate non-linearities for each module, with independently applied skip connections.

However, GECO does not precisely align with the taxonomy defined in the survey~\citep{min2022transformer}. 
In GECO, we did not just use LCB as an auxiliary module to Transformer.
Instead, we designed a new compact layer comprising local and global blocks.
We eliminated the intermediate non-linearities and parameters to reduce the overall number of parameters,
simplifying the training process. We applied skip connections to the entire GECO layer as a whole,
rather than separately. These deliberate design choices distinguish GECO from the use of off-the-shelf methods.
Please also refer to \cref{appendix:gps_comparison} for further design details.

Furthermore, \cref{tab:gt_survey} highlight that \model achieves consistent superior predictive quality across all datasets.
Specifically, on arxiv and molphcha, GECO achieves significant relative improvements of up to 28.11\% and 11.23\%, respectively.
We also note that, PE and AT are orthogonal approaches to our work.

\subsection{Comparison with Graph-ViT/MLP-Mixer}
\label{appendix:mlp-mixer}
\begin{table}[H]
    \caption{Comparison of \model with Graph-ViT and Graph-MLP-Mixer}
    \centering
    \fontsize{8pt}{8pt}\selectfont
    \setlength\tabcolsep{4pt}
    \label{tab:mlp_mixer}
    \begin{tabular}{lcc}
    \toprule
    \multirow{2}{*}{\makecell{\textbf{Model}}} &\textbf{Peptides-func} &\textbf{Peptides-struct} \\
    \cmidrule{2-3}
     &\textbf{AP $\uparrow$} &\textbf{MAE $\downarrow$} \\
     \midrule
        Graph-MLP-Mixer   & 0.6970 $\pm$ 0.0080	& 0.2475 $\pm$  0.0015 \\
        Graph-ViT  & 0.6942 $\pm$  0.0075&	0.2449 $\pm$  0.0016 \\
        \model  &	0.6975 $\pm$  0.0025&	0.2464 $\pm$  0.0009 \\
    \bottomrule
    \end{tabular}
\end{table}

\citep{he2023generalization} proposed two models, namely Graph-ViT and Graph-MLP-Mixer that
are generalization to popular ViT~\citep{dosovitskiy2020image} and MLP-Mixer~\citep{tolstikhin2021mlp} architectures to graph structures focusing on graph-level tasks,
such as those used in \cref{sec:pred_quality}.

In \cref{tab:mlp_mixer}, we summarize the results of \model, Graph-ViT, and Graph-MLP-Mixer where results are available across all works.
The results indicate that all models achieve competitive results and fall into each other’s confidence intervals, with \model achieving a better mean of Peptides-func, Graph-Vit achieving a better mean on Peptides struct.

However
Graph-ViT/MLP-Mixer only focuses on graph level tasks consisting of small number of average number,
and their scalability to the large node prediction datasets has no evidence.
This is likely because they require graph re-partitioning at every training epoch
using METIS partitioner.
While this approaches may be feasible for small graphs,
it is very costly strategy for large node prediction datasets such as those in \cref{sec:scalability_experiments}.
On the other hand, our method does not require such partitioning;
it can be simply used as a drop-in replacement for the self-attention of GTs.

\subsection{Comparison with GraphGPS}
\label{appendix:gps_comparison}

\definecolor{weightcolor}{RGB}{178,171,210}
\begin{figure}
  \begin{subfigure}{0.48\linewidth}
    \centering
    \includegraphics[width=\linewidth]{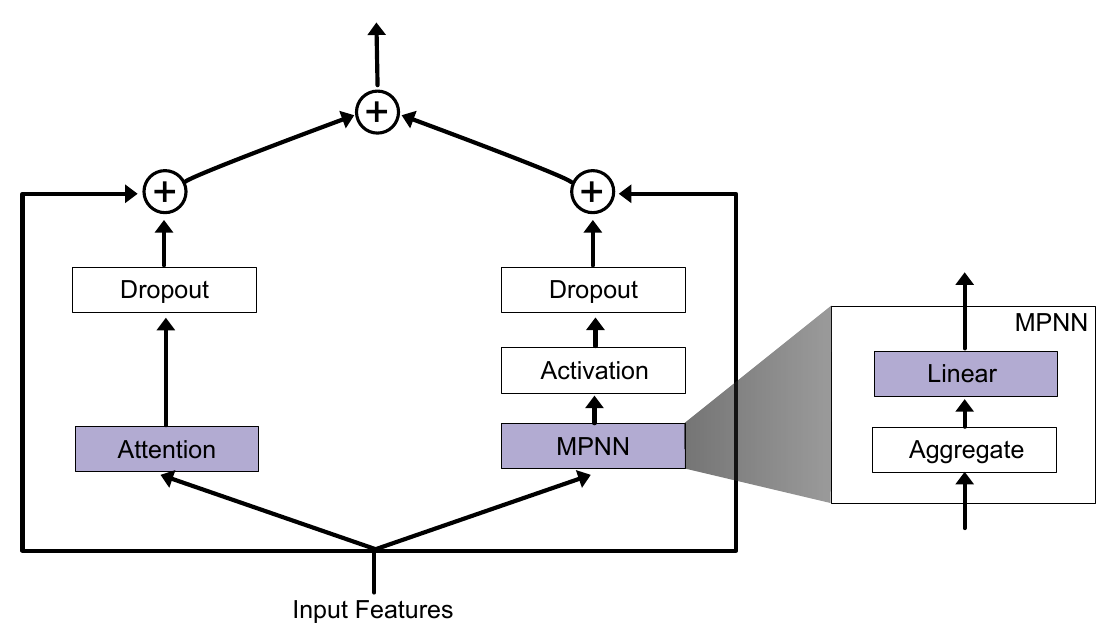}
    \caption{
    \textbf{GraphGPS Layer} consists of an MPNN and global attention modules with each module having its own skip connections and optional dropout.
    The modules are evaluated in parallel and summed up at the end.
    Both MPNN and attention modules usually have learnable weights.
    }
  \end{subfigure}
  \hspace{0.04\linewidth}
  \begin{subfigure}{0.48\linewidth}
    \centering
    \includegraphics[width=\linewidth]{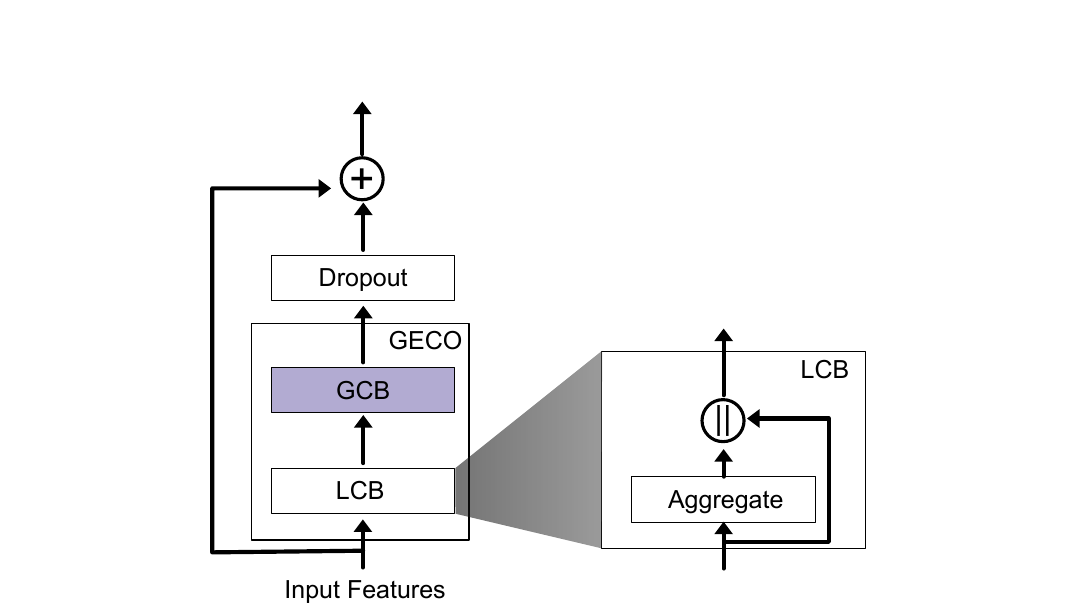}
    \caption{
    The \textbf{\model Layer} comprises \lcb (\lcba) and \gcb (\gcba), evaluated sequentially with a skip connection across the entire block.
    Notably, \lcba lacks learnable weights, serving as a pre-step to \gcba, which incorporates the learnable weights.
    }
  \end{subfigure}
  \caption{A comparison between GraphGPS and GECO, where the layers with learnable weights are highlighted in \textcolor{weightcolor}{color}.}
  \label{fig:gps_geco_comp}
\end{figure}

As \model can be considered a hybrid method, a natural question arises regarding what sets it apart from GraphGPS.
In this section, we delve into the distinctions between these two models.
The fundamental difference between \model and GraphGPS layers lies in their design as illustrated in Figure~\ref{fig:gps_geco_comp}.
Given input features, the MPNN and attention modules of GraphGPS are evaluated in parallel and then summed up.
Each module encompasses its own set of learnable weights, activation functions, dropout layers, and residual connections.
On the other hand, in the case of \model, \lcba and \gcba are evaluated sequentially.
Notably, \lcba does not incorporate any learnable weights, activation functions, or dropout layers;
it functions as a feature propagation block with a dense skip connection.
Instead, \model's weights are encapsulated within the \gcba block.
Furthermore, the residual connection is applied to the entire \model block.
In practice, \model and GraphGPS can be combined in various ways.
In our experiments, we chose to employ \model as an attention module to facilitate a direct comparison with GraphGPS and Exphormer.
However, it is possible to integrate GraphGPS and \model differently.
Potential options also include substituting \gcba with a Transformer or replacing \lcba with an MPNN.

\subsection{Connection between Jumping Knowledge Networks and NAGphormer}
\label{appendix:nagphormer}

\begin{figure}[h]
    \centering
    \includegraphics[width=0.2\textwidth]{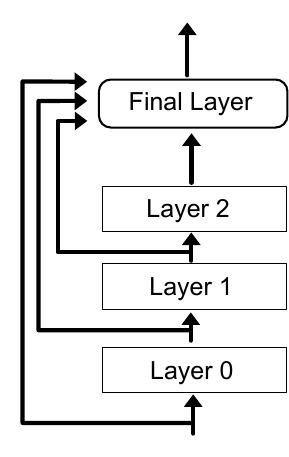}
    \caption{\label{fig:jknet}
            Illustration of JK-Nets with 3 layers.
            It is important to note that the Final Layer can be implemented using different layers,
            and it does not necessarily have to be the same as the intermediate layers. Although the original work 
            While the original work~\citep{xu2018representation} did not introduce a dense skip connection from the original inputs to the Final Layer, we have included it here for the sake of consistency in notation.
    }
\end{figure}

Jumping Knowledge Networks (JK-Nets)~\citep{xu2018representation} have been introduced as GNNs with a variant of dense skip connection.
The difference from the original dense skip connections~\citep{huang2017densely} is that instead of establishing dense connections between every
consecutive layer, JK-Nets establish dense skip connections from each layer to a final aggregation layer as illustrated in Figure~\ref{fig:jknet}.
Given this framework, we can recover NAGphormer as a special case of JK-Nets with two simple configurations:
\begin{enumerate}
    \item Replace non-linear transformation function of the GNN with the identity function. That is we do not use learnable weights, simply utilize traditional feature propagation.
    \item Set Final-Layer as multi-head attention.
\end{enumerate}
With these two simple modifications, one can recover the NAGphormer as a specific instance of JK-Nets.
Here NAGphormer corresponds to an MHA layer with tokens produced by a GNN with no learnable weights, or traditional feature propagation.

\end{document}